\documentclass[11pt]{article}

\usepackage[final]{acl}
\usepackage{tcolorbox}
\usepackage{times}
\usepackage{latexsym}
\usepackage[T1]{fontenc}
\usepackage[utf8]{inputenc}
\usepackage{microtype}
\usepackage{hyperref}
\usepackage{url}
\usepackage{booktabs}
\usepackage{xspace}
\usepackage{graphicx}
\usepackage{textcomp}
\usepackage{multicol}
\usepackage{multirow}
\usepackage{enumitem}
\usepackage{pifont}
\usepackage{amsmath}
\usepackage{subfigure}
\usepackage{float}
\usepackage{tikz}
\usepackage{subcaption}
\usepackage{lineno}
\usepackage{makecell}
\usepackage{CJKutf8}
\usepackage{inconsolata}
\usepackage{adjustbox}

\usepackage{placeins}

\definecolor{darkblue}{rgb}{0, 0, 0.5}
\hypersetup{colorlinks=true, citecolor=darkblue, linkcolor=darkblue, urlcolor=darkblue}

\definecolor{mycolor}{RGB}{33, 95, 154}
\definecolor{custom_red}{RGB}{228, 54, 54}
\def \name{\textsc{InterPol}\xspace}
\usepackage[table]{xcolor}
\usepackage{amssymb}

\usepackage{diagbox}
\title{InterPol: De-anonymizing LM Arena via Interpolated Preference Learning}

\author{}
\author{
  Minsung Cho \qquad Jaehyung Kim \\
  Yonsei University \\
  \texttt{\{mscho2008, jaehyungk\}@yonsei.ac.kr}
}

\begin{document}
\maketitle

\begin{abstract}

Voting-based leaderboards such as LM Arena are central to open-ended LLM evaluation, but their reliability depends on response anonymity.
Prior work shows that surface features such as bag-of-words and TF-IDF can reveal model identity from the response, yet such signals are unreliable when models are stylistically similar or closely matched, and are easily down-weighted during vote aggregation.
To examine the remaining anonymity risk, we introduce \name{}, a model-driven identification framework that learns target-specific response signatures from interpolated preference data. 
\name{} synthesizes controllable hard negatives by interpolating a target-mimicking copy model with its initial backbone, and trains a target-specific detector via curriculum-based triplet preference learning. 
Across three frontier target LLMs and two query sources, \name{} yields average relative gains of 32\% in Accuracy and 25\% in AUROC over prior methods.
Moreover, in arena-style promotion simulations, \name{} attains the same ranking gain with about 19\% fewer interactions than existing identification baselines; with competitor-aware voting, interactions are further reduced by about 69.9\% relative to target-only voting.

\end{abstract}
\section{Introduction}

The rapid emergence of Large Language Models (LLMs) \citep{touvron2023llama,achiam2023gpt,team2023gemini} has made their fair and reliable evaluation a central challenge. 
Conventional answer-based benchmarks \citep{chen2021evaluating, hendrycks2020measuring} neither reflect the diversity of real-world usage~\citep{liang2022holistic} nor fairly assess open-ended tasks that admit multiple valid responses~\citep{zheng2023judging}, driving a shift toward evaluation pipeline grounded in user preferences \citep{stiennon2020learning, zhou2023lima, dubois2024length}. 
One of the most representative examples of this paradigm is voting-based leaderboard such as \textit{LM Arena}~\citep{zheng2023judging, chiang2024chatbot}: users are shown anonymous responses from two LLMs to the same prompt, and their judgments are aggregated into scores such as Elo ratings \citep{boubdir2024elo}. 
The core assumption here is the \textit{anonymity of model responses}: each judgment is presumed to depend solely on response quality rather than the identity of the source model \citep{panickssery2024llm}. 
If this assumption is violated through identification signals embedded in the responses, coordinated manipulation targeting a specific LLM becomes feasible, undermining the credibility of the leaderboard \citep{singh2025leaderboard}.

Therefore, investigating the feasibility of LLM identification is critical to the reliability of evaluation with leaderboard, yet this line of work has only begun to be explored \citep{min2025improving, huang2025exploring}.
For instance, \citet{huang2025exploring} show that even simple statistical features, such as Bag-of-Words \citep{Salton1975AVS} or TF--IDF representations \citep{Salton1988TermWeightingAI}, can identify LLMs with high accuracy.
However, manipulations built on such surface-level features can be systematically mitigated; 
for example, defenders can down-weight suspicious votes by exploiting the same statistical signals during leaderboard aggregation \citep{lmarena2024sentiment, lmarena2024style}. 
While this defensibility alleviates some concerns, it raises a key question: \textit{do more advanced de-anonymization (i.e., identification) techniques exist that could threaten the robustness and credibility of leaderboards?}

\begin{figure*}[t]
\vspace{-0.1in}
  \centering
  \begin{minipage}[t]{0.50\linewidth}
    \centering
    \includegraphics[width=\linewidth]{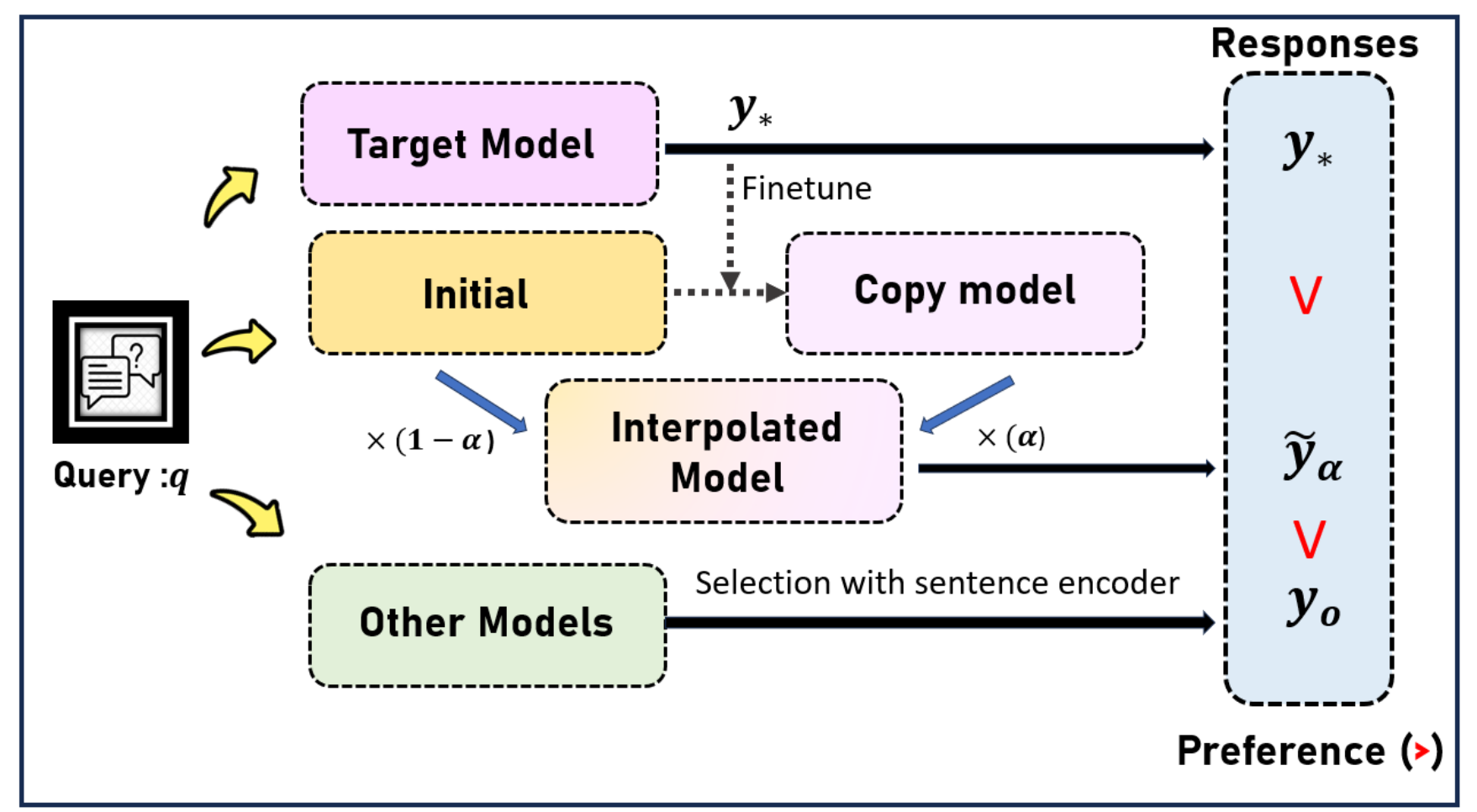}
    \\(a) Negative synthesis via model interpolation
  \end{minipage}
  \hfill
  \hfill
  \begin{minipage}[t]{0.48\linewidth}
    \centering
    \includegraphics[width=\linewidth]{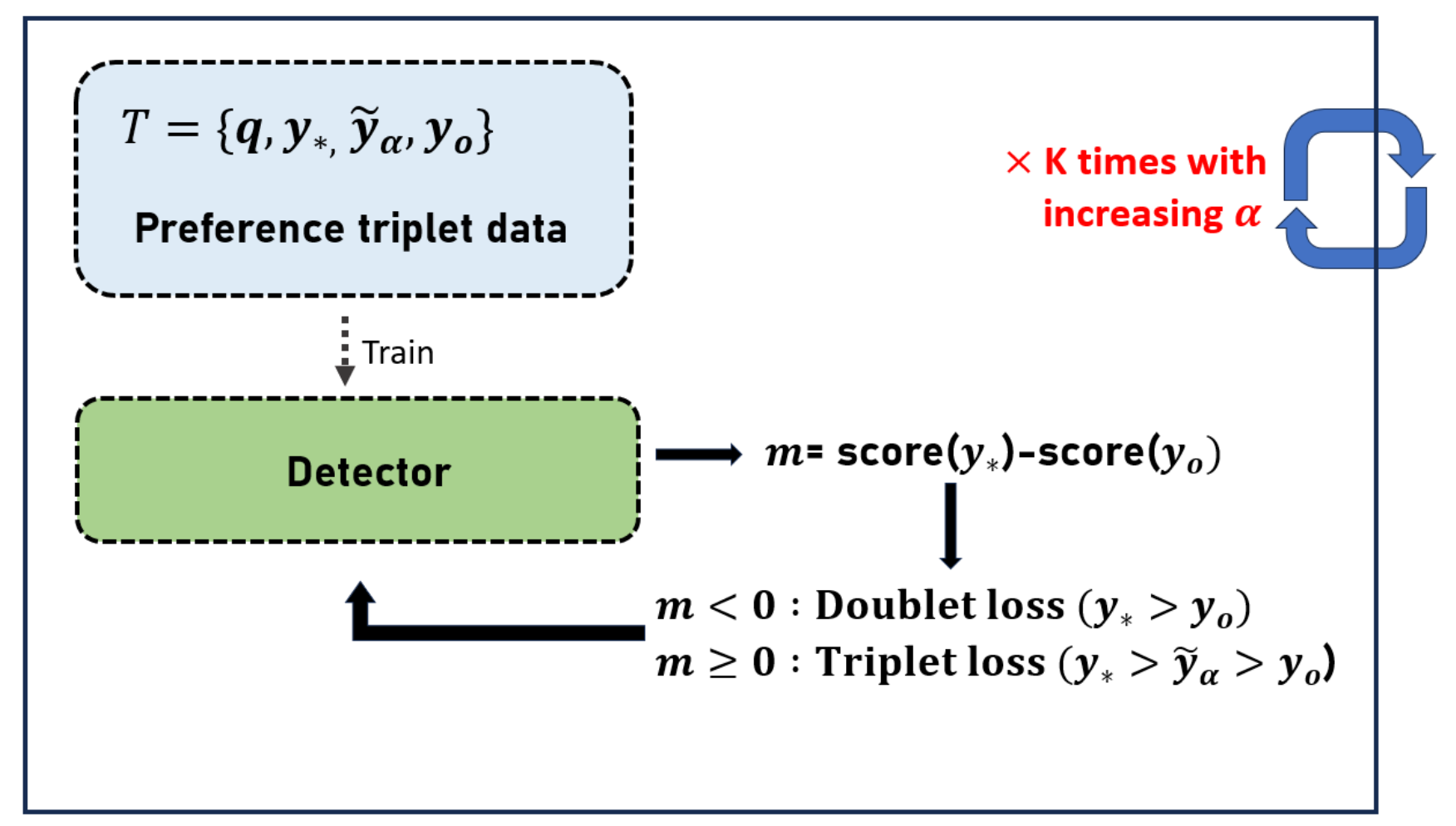}
    \\(b) Adaptive curriculum learning
  \end{minipage}

\caption{
\textbf{Overview of \name{}.}
(a) Target-mimicking copy model is interpolated with its initial backbone to generate synthetic hard negatives with controllable similarity (\textit{i.e.}, difficulty) to the target LLM. 
These responses form training triplets together with the target response and the most similar non-target response.
(b) Detector uses pairwise or triplet preference objectives according to the margin between target and non-target responses, while increasing $\alpha$ across $K$ stages to progressively raise the task difficulty.
}

\label{fig:main_figure}
\vspace{-0.1in}
\end{figure*}

In this paper, we answer this question affirmatively by proposing \textbf{\name{}}, a model-driven LLM identification framework that learns from \textbf{InterPol}ated preference data.
Our key idea is to train a target-specific detector that scores a target LLM's responses higher (\textit{i.e.,} prefer) than other LLMs', using only queries and responses collected from multiple LLMs without metadata or declared identity.
However, responses naively collected from other LLMs are often easy to distinguish, providing limited signal for training a strong detector. 
To this end, we synthesize hard negatives via \textit{model interpolation} \citep{wortsman2022model, ilharco2022editing}: we first fine-tune a separate copy model to mimic the target LLM, then construct an interpolated model between this fine-tuned model and the base model. 
\name{} uses the responses from the interpolated model as hard negatives for preference learning, where the interpolation ratio serves as an explicit control over their difficulty.
With this data, \name{} trains the detector on the inherent preference order within triplets of $\langle$target, interpolated, other LLMs$\rangle$ responses, using adaptive curriculum learning that gradually increases the interpolation ratio to expose the detector to increasingly harder negatives (see Figure~\ref{fig:main_figure}). 

To evaluate \name{}, we first conduct target LLM identification experiments across three frontier target LLMs (GPT-4o, Sonnet-4, and Gemini-2.5-Pro) and two query sources (Section~\ref{sec:Experiment_setup}).
Across these settings, \name{} consistently outperforms six representative baselines in both identification accuracy and AUROC; specifically, relative to the average of prior baselines, \name{} improves accuracy/AUROC by 32.3\%/25.0\%, respectively.
Its advantage also persists in more challenging setups of close-Elo and within-family comparisons, where coarse capability gaps, model-scale differences, and surface-level statistical cues are less informative for attribution. 

We then connect this identifiability to its practical impact through ranking manipulation simulations on Arena-style battle data (Section~\ref{sec:5}). 
Using reconstructed leaderboard dynamics, we quantify the number of adversarial votes and total interactions required to move target models across ranks.
Across representative promotion settings, \name{} reduces the required interaction budget by 19\% on average compared to existing identification baselines. 
Moreover, when voting strategy includes downvotes against competitors, required interaction further decreases by about 69.9\% compared with a target-only voting strategy. 
Overall, our results show that response identifiability is not merely an offline attribution signal, but a concrete reliability risk that can lower the cost of strategic voting and alter leaderboard outcomes.

\noindent In summary, our contributions are threefold.
\begin{itemize}[leftmargin=*, itemsep=0pt, topsep=2pt]
\item[$\circ$]  We reveal an anonymity risk beyond surface cues: anonymous LLM responses retain target-specific signatures that enable source attribution.
\item[$\circ$] We propose \name{}, which learns these signatures using interpolated hard negatives and triplet preference learning with an adaptive curriculum.
\item[$\circ$] We demonstrate that this identifiability is robust across query sources and harder competitor settings, and that it reduces the interaction cost of strategic voting in Arena-style simulations.
\end{itemize}

\section{Related Works}


\paragraph{LLM evaluation and preference leaderboards.}
Early LLM evaluation relied on static benchmarks such as GLUE~\citep{wang2018glue}, HumanEval~\citep{chen2021evaluating}, and MMLU~\citep{hendrycks2020measuring}.
While these offer controlled and reproducible measurement, they struggle to capture open-ended generation and diverse real-world use~\citep{liang2022holistic}, and are increasingly susceptible to test data contamination~\citep{magar2022data}.
Recent evaluation has therefore shifted toward preference-based protocols, including LLM-as-a-judge methods~\citep{zheng2023judging} and large-scale human preference platforms.
LM Arena is a prominent example: users compare two anonymized model responses to the same prompt, and the resulting pairwise preferences are aggregated into continuously updated leaderboard ratings~\citep{chiang2024chatbot,boubdir2024elo}.
This evaluation paradigm makes response anonymity central to leaderboard reliability.


\paragraph{De-anonymization and leaderboard vulnerabilities.}
The anonymity of pairwise comparisons can fail when model identity is inferable from generated responses, enabling source attribution and downstream ranking manipulation~\citep{min2025improving,huang2025exploring,zhao2025challenges}.
Such identifiability typically arises from emergent statistical regularities in model outputs, \textit{e.g.}, BoW or TF--IDF signatures, and is closely related to the broader problem of attributing text to its source model \citep{uchendu2021turingbench, antoun2024text}.
Also, this is distinct from approaches that insert identity signals deliberately, such as watermarking and fingerprinting~\citep{kirchenbauer2023watermark,xu2024instructional}.
However, emergent statistical cues become unreliable when models are stylistically similar, closely ranked, or subject to preference-bias corrections.
To our knowledge, we are the first to cast leaderboard de-anonymization as a learning problem, training a target-specific detector rather than relying on hand-crafted surface features.

\section{\name{}: Interpolated Preference Learning for LLM Identification}

\paragraph{Problem setup.} 
Let us denote a set of accessible LLMs as $\mathcal{M} = \{M_1, M_2, \ldots, M_K\}$ where each LLM $M \in \mathcal{M}$ can generate response $y$ for the given input query $q$, \textit{i.e.}, $y \sim M(q)$. 
Then, our goal de-anonymizing the leaderboard (\textit{i.e.}, identify source LLM) is to construct a detector model $f_{\theta}$ that assigns higher scores to response $y_{*}$ generated by a target model $M_*$ compared to response $y_k$ from other models $M_k \in \mathcal{M} \setminus \{M_*\}$:
\begin{equation}
f_{\theta}(q, y_{*}) > f_{\theta}(q, y_k), \quad \forall M_k \in \mathcal{M} \setminus \{M_*\}.
\end{equation}
We assume that we have a set of queries $\mathcal{Q} = \{q^i\}_{i=1}^{N}$ and a set of responses $\mathcal{Y}=\{y^{i}_{k}|y^{i}_{k}\sim M_{k}(q^i),q^i \in Q, M_k \in \mathcal{M}\}$.
Here, $\mathcal{Y}$ can be constructed by querying each model (\textit{e.g.}, public API), without leaderboard access.
Then, one can learn the detector model $f_{\theta}$ using $\mathcal{Q}$ and $\mathcal{Y}$; for instance, a simple regression model based on bag-of-words \citep{huang2025exploring}. 
However, these surface-level features often fail to capture the subtle stylistic and reasoning patterns that distinguish modern LLMs.
Therefore, \name{} instead trains the detector by finetuning LLM on $\mathcal{Q}$ and $\mathcal{Y}$.

\paragraph{Threat model.}
We consider an attacker who is a model developer seeking to raise the ranking of their own model on a public, voting-based leaderboard such as LM Arena. 
The attacker participates as a normal user without access to system internals. 

As shown in prior works  \citep{zhao2025challenges,huang2025exploring}, strategic voting alone is sufficient to manipulate leaderboard outcomes. 
A crucial requirement for such manipulation is the ability to de-anonymize model responses, \textit{i.e.}, to infer which competing LLM produced each answer in a duel so that the attacker can cast selective votes that most benefit their target model. 
Prior to deployment, the attacker collects model responses by querying publicly accessible models through their APIs, including the target model, and uses the collected responses to train a target-specific detector.
At test time, the trained detector including \name{} enables the attacker to identify the target model's responses among previously unseen anonymous responses encountered on the leaderboard, using only the response text and without access to model identity labels, leaderboard metadata, or privileged platform information.

\subsection{Identification via Preference Learning}\label{sec:3.1}

To train the detection model, we first construct a dataset of preference pairs:
\begin{equation*}
\mathcal{D} = \{(q^i, y_*^i, y_o^i)\}_{i=1}^N,
\end{equation*}
where for each query $q^i \in \mathcal{Q}$, $y_*^i$ represents the response from target model $M_*$, and  $y_o^i$ represents the selected responses from other models $\mathcal{Y}^{i}=\{y^{i}_k|y^{i}_{k}\sim M_{k}(q^i),\forall M_k \in \mathcal{M} \setminus \{M_*\}\}$.
To maximize discrimination difficulty during training, we select a hard negative $y^{i}_k$ as $y_o^i$ by computing semantic similarity between the target response, and selecting the most similar one:
\begin{equation}\label{eq:neg}
y_o^i = \arg\max_{y_k^i \in \mathcal{Y}^i}~\text{sim}\big(\mathbf{e}(y_*^i), \mathbf{e}(y_k^i)\big),
\end{equation}
where $\mathbf{e}(\cdot)$ represents embeddings from the sentence encoder, and $\text{sim}(\cdot, \cdot)$ is a cosine similarity.
Then, the detector model $f_{\theta}$ is trained to prefer target response  $y_*^i$ over others' response $y_k^i$ by minimizing the following binary cross-entropy loss:
\begin{align}
\mathcal{L}_{\mathrm{pair}}
&= \frac{1}{|\mathcal{D}|}
\sum_{i=1}^{|\mathcal{D}|}
\mathcal{L}_{\mathrm{pref}}(q^i, y_*^i, y_o^i),
\label{eq:pair} \\
\mathcal{L}_{\mathrm{pref}}(q,y_w,y_l)
&= -\log \sigma\!\left(
f_{\theta}(q, y_w) - f_{\theta}(q, y_l)
\right).
\nonumber
\end{align}
This loss function enforces the preference modeling based on Bradley-Terry (BT) model \citep{bradley1952rank} by maximizing the probability that the target response receives a higher likelihood than the competing response.

For training, we use only the single hardest negative $y_o^i$. 
However, during validation and test, we expand the evaluation to include responses from all non-target LLMs. 
Namely, given $K$ LLMs, we evaluate against the remaining $K-1$ LLMs (excluding the target) to comprehensively assess detector performance across the entire model space.

\subsection{Hard Negative Synthesis via Distillation and Model Interpolation}\label{sec:3.2}

The promising way to improve the performance of detector $f_{\theta}$ is enlarging the size of $\mathcal{D}$ (\textit{i.e.}, larger $N$) or expanding the diversity of $y_o^i$ by incorporating more LLMs (\textit{i.e.}, larger $K$).
However, this approach often incurs the additional costs for new data collection such as increased API callings. 
This motivates our approach: \textit{Can we generate synthetic hard negatives while maintaining high discrimination difficulty?} 
To answer this question, we propose a new approach that synthesizes hard negatives by leveraging model interpolation \citep{wortsman2022model, ilharco2022editing}.

Specifically, the proposed approach begins by creating \textit{copy model} $\hat{M}_*$ that is trained to mimic the response of target model ${M}_*$ through supervised fine-tuning (\textit{i.e.}, off-policy knowledge distillation):
\begin{equation}
\hat{M}_* = \arg\min_{\phi} \frac{1}{|\mathcal{D}|} \sum_{(q^i, y_*^i) \in \mathcal{D}} \mathcal{L}_{CE}(\hat{M}_{\phi}(q^i), y_*^i),
\end{equation}
where $\mathcal{L}_{CE}$ denotes a standard language modeling loss. 
After training, $\hat{M}_*$ serves as the foundation for generating synthetic responses by varying the levels of similarity to the target model ${M}_*$: 
 we synthesize responses of intermediate-difficulty from the model $\widetilde{M}_{\alpha}$ which is constructed with an interpolated model parameter $\widetilde{\phi}_{\alpha}$:
\begin{equation}
\widetilde{\phi}_{\alpha} = (1-\alpha) \phi_{\tt init} + \alpha \hat{\phi}_{\tt copy},
\end{equation}
where $\alpha\in[0,1]$ is hyperparameter to control difficulty, $\phi_{\tt init},\hat{\phi}_{\tt copy}$ denote the parameters of initial copy model and trained model $\hat{M}_*$, respectively. 

Then, using $\widetilde{M}_{\alpha}$, we construct the response-triplet dataset as below:
\begin{equation}
\mathcal{T}_{\alpha} = \{(q^i, y_*^i, \widetilde{y}_{\alpha}^i, y_o^i)\}_{i=1}^N,
\end{equation}
where $\widetilde{y}_{\alpha}^i = \widetilde{M}_{\alpha}(q^i)$  is the response from interpolated model. 
Because weight-space interpolation induces smooth functional interpolation, the generations of $\widetilde{M}_\alpha$ vary between the initializer and the copy model: as $\alpha\!\to\!0$ they resemble the initializer (\textit{easy-to-discriminate}), as $\alpha\!\to\!1$ they approach the target-like copy (\textit{hard-to-discriminate}), and for $\alpha\!\in(0,1)$ they yield {middle-difficulty negatives}, typically satisfying
\[
0 < \Delta(q,\widetilde{y}_\alpha) < \Delta(q,y_o),
\]
where $\Delta(q,y) = f_\theta(q,y_*) - f_\theta(q,y)$.

\subsection{Adaptive Curriculum Learning}\label{sec:3.3}

With the constructed response-triplet dataset $\mathcal{T}_{\alpha}$, we train our detector model $f_{\theta}$ by minimizing the following linear combination of binary cross-entropy losses:
\allowdisplaybreaks
\begin{align}
\mathcal{L}_{\tt fin}
&= \frac{1}{|\mathcal{T}_{\alpha}|}
   \sum_{i \in \mathcal{T}_{\alpha}} \mathcal{L}^i_{\tt trp}, \label{eq:triple} \\
\mathcal{L}^i_{\tt trp}
&= \lambda_1 \mathcal{L}_{\tt pref}(q^i, y^i_*, \widetilde{y}^i_{\alpha}) \nonumber \\
&\quad + \lambda_2 \mathcal{L}_{\tt pref}(q^i, \widetilde{y}^i_{\alpha}, y^i_o) \nonumber \\
&\quad + \lambda_3 \mathcal{L}_{\tt pref}(q^i, y^i_*, y^i_o). \nonumber
\end{align}
where $\lambda_1, \lambda_2, \lambda_3$ are hyperparameters that weigh the relative importance of each triplet component.
While this triplet training is effective, some queries might suffer during training if they even fail to minimize a simple pair-wise loss (Eq.~\ref{eq:pair}).
To improve training by mitigating this issue, we consider adaptive learning.
Specifically, we first introduce discrimination margin:
\begin{equation}\label{eq:adaptive}
    m^i = f_{\theta}(q^i, y_*^i) - f_{\theta}(q^i, y_o^i),
\end{equation}
where $m^i<0$ indicates that the detector $f_\theta$ currently fails to correctly discriminate target response $y_*^i$ and the hard negative response $y_o^i$. 
For those samples with $m^i<0$, we then apply $\mathcal{L}_{\tt pref}$ instead of $\mathcal{L}_{\tt trp}$ and focus on solving easier problem.

In addition, to further improve the performance of detection model $f_{\theta}$, we introduce curriculum learning that progressively increase discrimination difficulty by varying the degree of interpolation $\alpha$.
Specifically, we initially construct easier triplet data $\mathcal{T}_{\alpha_1}$ with small value of $\alpha_1$ and training the detection model following Eq.~\ref{eq:triple}. 
After this stage, we construct a harder dataset $\mathcal{T}_{\alpha_2}$ with a larger value $\alpha_2 > \alpha_1$ and continue training. 
In this way, we perform iterative data construction and training across progressively harder stages, where we set $K=2$ (with $\alpha_1=0.5,\alpha_2=0.75$) in our experiments.

\section{Experiment}
\label{sec:Experiment_setup}

\paragraph{Datasets and candidate pools.}
We evaluate \name{} on two query sources: the \textit{Alpaca dataset}~\citep{alpaca} and the \textit{Arena human preference dataset}~\citep{tang2025arena_explorer}.
We collect responses from 12 LLMs spanning six model families and split the data into training, validation, and test sets.
We consider three primary target LLMs: \textit{GPT-4o}~\citep{gpt4o_report}, \textit{Gemini-Pro}~\citep{comanici2025gemini25pushingfrontier}, and \textit{Claude-Sonnet-4}~\citep{claude4_system_card}.
In addition to the standard 12-model pool, we use a close-Elo competitor pool and a composed pool for generalization analysis. 
More details such as full dataset statistics and generation details are provided in Appendix~\ref{tab:datasets and metrics}.

\paragraph{Baselines.}
We consider six representative baselines.
(1) \textit{Length--Word} and (2) \textit{Length--Char} use response length measured by word count and character count, respectively.
(3) \textit{TF--IDF} \citep{Salton1988TermWeightingAI} and (4) \textit{Bag-of-Words} \citep{Salton1975AVS} use lexical $n$-gram features with logistic-regression classifiers.
(5) \textit{Neural-based multi-class classifier} \citep{tay2020reverse} trains a supervised source-attribution model to predict the generating LLM. 
(6) \textit{LLM-as-a-Judge} \citep{zheng2023judging} with few-shot in-context learning asks a judge model to identify which candidate response is more likely to come from the target LLM.
Implementation and evaluation details are in Appendix~\ref{app:baseline_details}.

\paragraph{Implementation.}
We use GRM-Llama-3.2-3B as the detector backbone and Phi-4-mini-instruct as the target-mimicking copy-model backbone.
The detector is trained with a two-stage curriculum using $\alpha=0.5$ and $0.75$.
Full implementation details are provided in Appendix~\ref{Training and implementation details}.

\subsection{Identification Results}
\label{sec:main_results}

\begin{table}[t]
\centering
\small

\caption{
\textbf{Main target-present identification results.}
We report average Accuracy (Acc.) and AUROC across 11 non-target LLMs on the Arena Human Preference and Alpaca datasets. 
Best scores are highlighted in \textbf{bold}.
Method abbreviations: Len-W = Length--Word, Len-C = Length--Char, TF-IDF = TF--IDF, BoW = Bag-of-Words, Neural = Neural-based classification, and Judge = LLM-as-a-Judge.
}
\vspace{-0.05in}
\label{tab:main_result}
\resizebox{\columnwidth}{!}{%
\begin{tabular}{@{}llcccc@{}}
\toprule
\multirow{2}{*}{\textbf{Target}} 
& \multirow{2}{*}{\textbf{Method}} 
& \multicolumn{2}{c}{\textbf{Arena}} 
& \multicolumn{2}{c}{\textbf{Alpaca}} \\
\cmidrule(lr){3-4} \cmidrule(lr){5-6}
& & Acc. & AUROC & Acc. & AUROC \\
\midrule

\multirow{7}{*}{GPT-4o}
& Len-W   & 0.556 & 0.611 & 0.493 & 0.551 \\
& Len-C   & 0.564 & 0.610 & 0.495 & 0.552 \\
& TF-IDF  & 0.697 & 0.766 & 0.655 & 0.704 \\
& BoW     & 0.709 & 0.772 & 0.631 & 0.692 \\
& Neural  & 0.898 & 0.888 & 0.726 & 0.767 \\
& Judge   & 0.822 & --    & 0.679 & --    \\
& \cellcolor{gray!20}\name{} 
& \cellcolor{gray!20}\textbf{0.908} 
& \cellcolor{gray!20}\textbf{0.920} 
& \cellcolor{gray!20}\textbf{0.887} 
& \cellcolor{gray!20}\textbf{0.875} \\
\midrule

\multirow{7}{*}{Gemini-Pro}
& Len-W   & 0.792 & 0.830 & 0.618 & 0.647 \\
& Len-C   & 0.768 & 0.815 & 0.618 & 0.639 \\
& TF-IDF  & 0.844 & 0.926 & 0.780 & 0.851 \\
& BoW     & 0.887 & 0.948 & 0.798 & 0.911 \\
& Neural  & 0.908 & 0.940 & 0.828 & 0.894 \\
& Judge   & 0.958 & --    & 0.620 & --    \\
& \cellcolor{gray!20}\name{} 
& \cellcolor{gray!20}\textbf{1.000} 
& \cellcolor{gray!20}\textbf{0.979} 
& \cellcolor{gray!20}\textbf{0.946} 
& \cellcolor{gray!20}\textbf{0.967} \\
\midrule

\multirow{7}{*}{Sonnet-4}
& Len-W   & 0.639 & 0.621 & 0.525 & 0.529 \\
& Len-C   & 0.380 & 0.397 & 0.519 & 0.523 \\
& TF-IDF  & 0.810 & 0.902 & 0.763 & 0.834 \\
& BoW     & 0.851 & 0.929 & 0.767 & 0.861 \\
& Neural  & 0.883 & 0.908 & 0.825 & 0.899 \\
& Judge   & 0.902 & --    & 0.631 & --    \\
& \cellcolor{gray!20}\name{} 
& \cellcolor{gray!20}\textbf{0.946} 
& \cellcolor{gray!20}\textbf{0.976} 
& \cellcolor{gray!20}\textbf{0.982} 
& \cellcolor{gray!20}\textbf{0.954} \\
\bottomrule
\end{tabular}
}
\vspace{-0.15in}
\end{table}

Table~\ref{tab:main_result} reports target-present identification results on both the Arena Human Preference and Alpaca datasets.
Each comparison contains one response from the target model and one response from a non-target model to the same query, and the detector selects the response assigned the higher target score.
Across both query sources, \name{} achieves the strongest overall performance for all three target LLMs.
On Arena, \name{} improves over the baseline average by 23.5\% in Accuracy and 21.2\% in AUROC.
On Alpaca, the corresponding gains are 41.1\% in Accuracy and 28.8\% in AUROC.
These improvements show that the advantage of \name{} is not confined to a single query distribution, but holds across both user-generated Arena-style queries and standard instruction-following queries.

The full model-wise Accuracy and AUROC breakdowns in 
Appendix Tables~\ref{tab:modelwise_human_accuracy} and~\ref{tab:modelwise_human_auroc}
further show that this trend also holds in within-family comparisons, including
GPT-4o vs. GPT-4o-mini, Gemini-Pro vs. Gemini-Flash, and Sonnet-4 vs. Sonnet-3.5.
Such comparisons are more challenging because models within the same family may share similar training pipelines, response styles, or distillation relationships.
While feature-based baselines become less reliable in these cases, \name{} remains more stable, indicating that interpolated preference learning captures target-specific response signatures beyond broad family-level or surface statistical cues.

\paragraph{Close-Elo competitor evaluation.}
The standard 12-model pool provides a broad evaluation of target-present identification, but it can contain models that differ substantially in family, scale, and capability.
To further reduce coarse separability among candidates, we evaluate \name{} on a close-Elo competitor pool consisting of high-ranked models clustered within a narrow range on the LM Arena text leaderboard. 
The pool includes our three target models, GPT-4o, Gemini-2.5-pro, and Claude-sonnet-4, together with five close competitors: Grok-4-Fast, DeepSeek-v3.1, GLM-4.6, Kimi-K2-0905, and Qwen3-Max-0923.
The Elo ratings and leaderboard snapshot date used to construct this pool are described in Appendix~\ref{app:close_elo_breakdown}.

The results in Tables~\ref{tab:close_elo_baseline}--\ref{tab:close_elo_modelwise_auroc} show that \name{} remains effective under this close-Elo evaluation. \name{} achieves the strongest performance for GPT-4o and Sonnet-4, and remains competitive for Gemini-Pro, where the Neural-Based baseline is slightly stronger.
Across all targets, \name{} consistently outperforms feature-based baselines, indicating that its gains are not driven solely by coarse separability in the standard pool but also persist among direct high-ranked competitors with similar leaderboard ratings.


\subsection{Additional Analyses}
\label{sec:analysis}

We further analyze what enables \name{} to identify target-model responses and how its training procedure improves discrimination.
We examine response-level attribution signals, component ablation, interpolation difficulty, score distributions, and interpretability. 
Further results and expanded evaluations are provided in Appendix~\ref{appendix:c}.

\paragraph{Response-level attribution signals.}
We first ask whether the attribution signal is present in the response itself, or whether the detector mainly relies on query-conditioned answer quality. 
If the detector only exploits how well a response matches a particular query, then removing the query should substantially weaken identification.
To test this, we evaluate a response-only variant that removes the query and scores each response independently.

\begin{table}[t]
\centering
\small
\caption{
\textbf{Response-only ablation.}
We compare the standard query-conditioned detector with a response-only variant that removes the query input.
}
\vspace{-0.05in}
\label{tab:response_only}
\resizebox{\columnwidth}{!}{%
\begin{tabular}{lcccc}
\toprule
\multirow{2}{*}{\textbf{Target}}
& \multicolumn{2}{c}{\textbf{Query-conditioned}} 
& \multicolumn{2}{c}{\textbf{Response-only}} \\
\cmidrule(lr){2-3} \cmidrule(lr){4-5}
& Accuracy & AUROC & Accuracy & AUROC \\
\midrule
GPT-4o      & 0.908 & 0.920 & 0.879 & 0.851 \\
Gemini-Pro  & 1.000 & 0.979 & 0.989 & 0.991 \\
Sonnet-4    & 0.946 & 0.976 & 0.981 & 0.983 \\
\bottomrule
\end{tabular}
}
\vspace{-0.1in}
\end{table}

As shown in Table~\ref{tab:response_only}, the performance is mostly preserved which indicate that substantial attribution signal is already present in the responses themselves.
For Gemini-Pro and Sonnet-4, the response-only variant performs comparably to, or even better than, the query-conditioned detector.
Specifically, response-only evaluation achieves 0.989 Accuracy / 0.991 AUROC for Gemini-Pro and 0.981 Accuracy / 0.983 AUROC for Sonnet-4.
For GPT-4o, response-only performance decreases moderately, from 0.908 to 0.879 Accuracy and from 0.920 to 0.851 AUROC, but remains clearly effective.
These results show that \name{} is not merely exploiting query-conditioned answer quality. 
Rather, anonymous responses themselves contain model-specific behavioral signatures, directly supporting the vulnerability of response anonymity in Arena-style evaluations.

\paragraph{Component ablation.}
\begin{table}[t]
\centering
\small


\caption{\textbf{Component ablation.} Comparison of different configurations of \name{} to train detector model.
We vary the use of negative sampling strategy (Eq.~\ref{eq:neg}), triplet loss (Eq.~\ref{eq:triple}),
adaptive curriculum (Eq.~\ref{eq:adaptive}), and iterative training ($K=1$ or $2$).
For compactness, we denote these components as \textbf{N} (Negative), \textbf{T} (Triplet),
\textbf{A} (Adaptive), and \textbf{I} (Iterative).
We report average Accuracy and AUROC across three target LLMs.
The best scores are highlighted in \textbf{bold}.}
\vspace{-0.05in}
\begin{tabular}{c c c c c c}
\toprule
\textbf{N} & \textbf{T} & \textbf{I} & \textbf{A} & {Accuracy} & {AUROC} \\
\midrule
Hard   & -- & -- & -- & 0.863 & 0.882 \\
Hard   & \checkmark & -- & -- & 0.914 & 0.931 \\
Hard   & \checkmark & -- & \checkmark & 0.935 & 0.951 \\
Easy   & \checkmark & -- & \checkmark & 0.887 & 0.896 \\
Random & \checkmark & -- & \checkmark & 0.928 & 0.920 \\
\rowcolor{gray!40}
Hard (Ours)   & \checkmark & \checkmark & \checkmark 
& \textbf{0.951} & \textbf{0.958} \\
\bottomrule
\end{tabular}
\label{tab:ablation_study}
\vspace{-0.15in}
\end{table}

\begin{figure*}[t]
    \vspace{-0.1in}
  \centering
  \begin{minipage}[t]{0.32\linewidth}
    \centering
    \includegraphics[width=\linewidth]{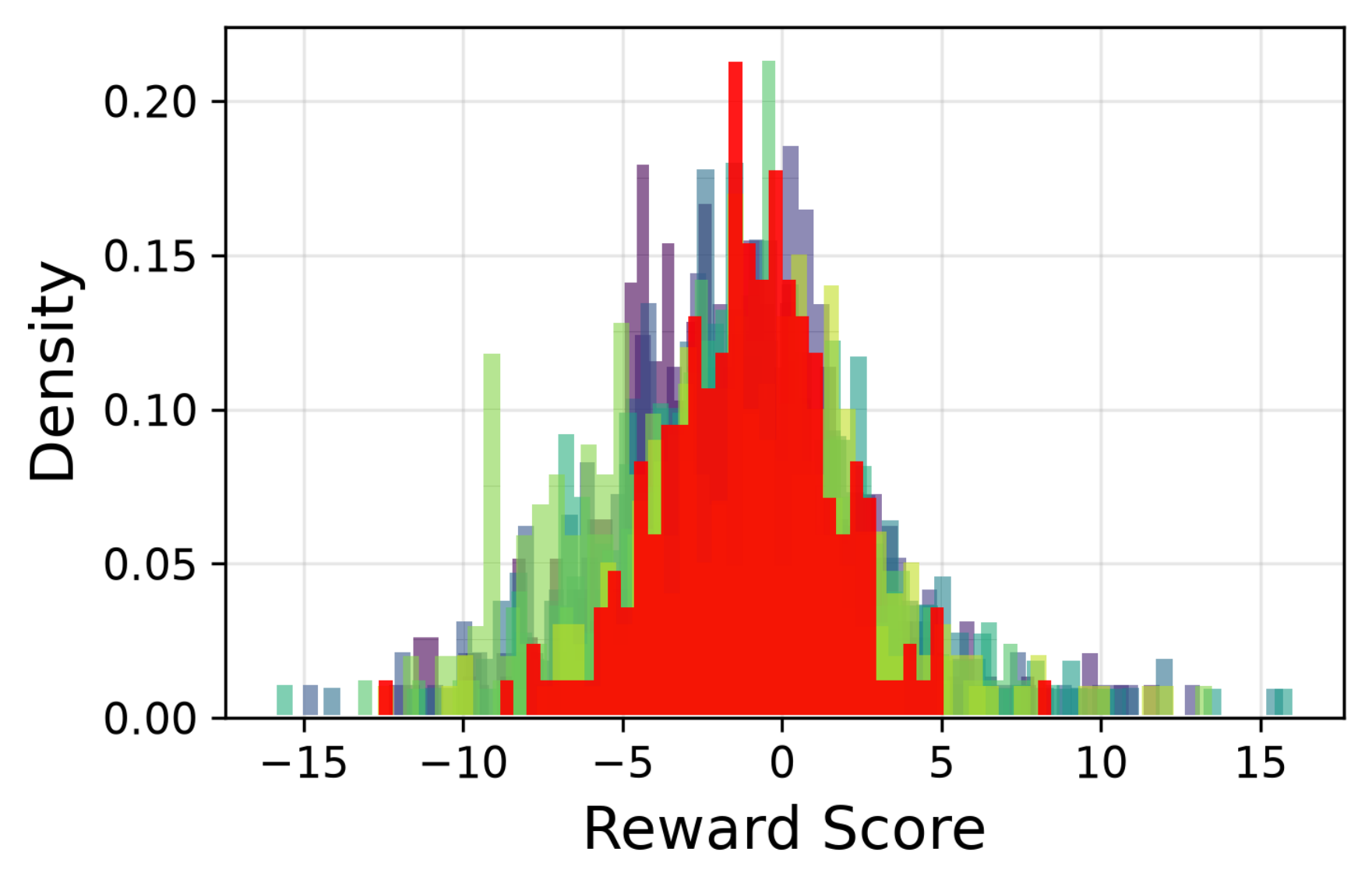}
    \\(a) Initial detector
  \end{minipage}
  \hfill
  \begin{minipage}[t]{0.32\linewidth}
    \centering
    \includegraphics[width=\linewidth]{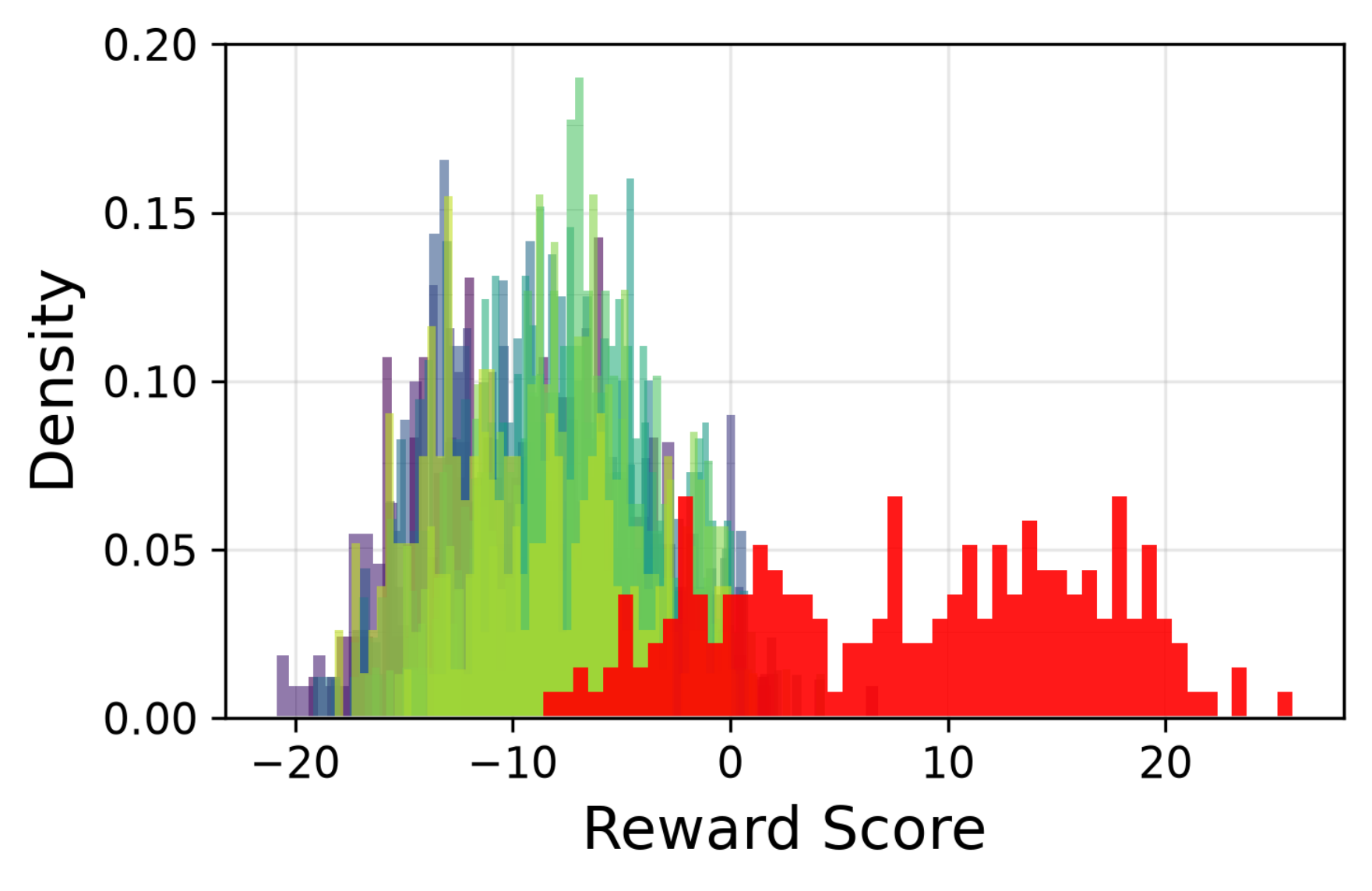}
    \\(b) After 1st stage ($\alpha=0.5$)
  \end{minipage}
  \hfill
  \begin{minipage}[t]{0.32\linewidth}
    \centering
    \includegraphics[width=\linewidth]{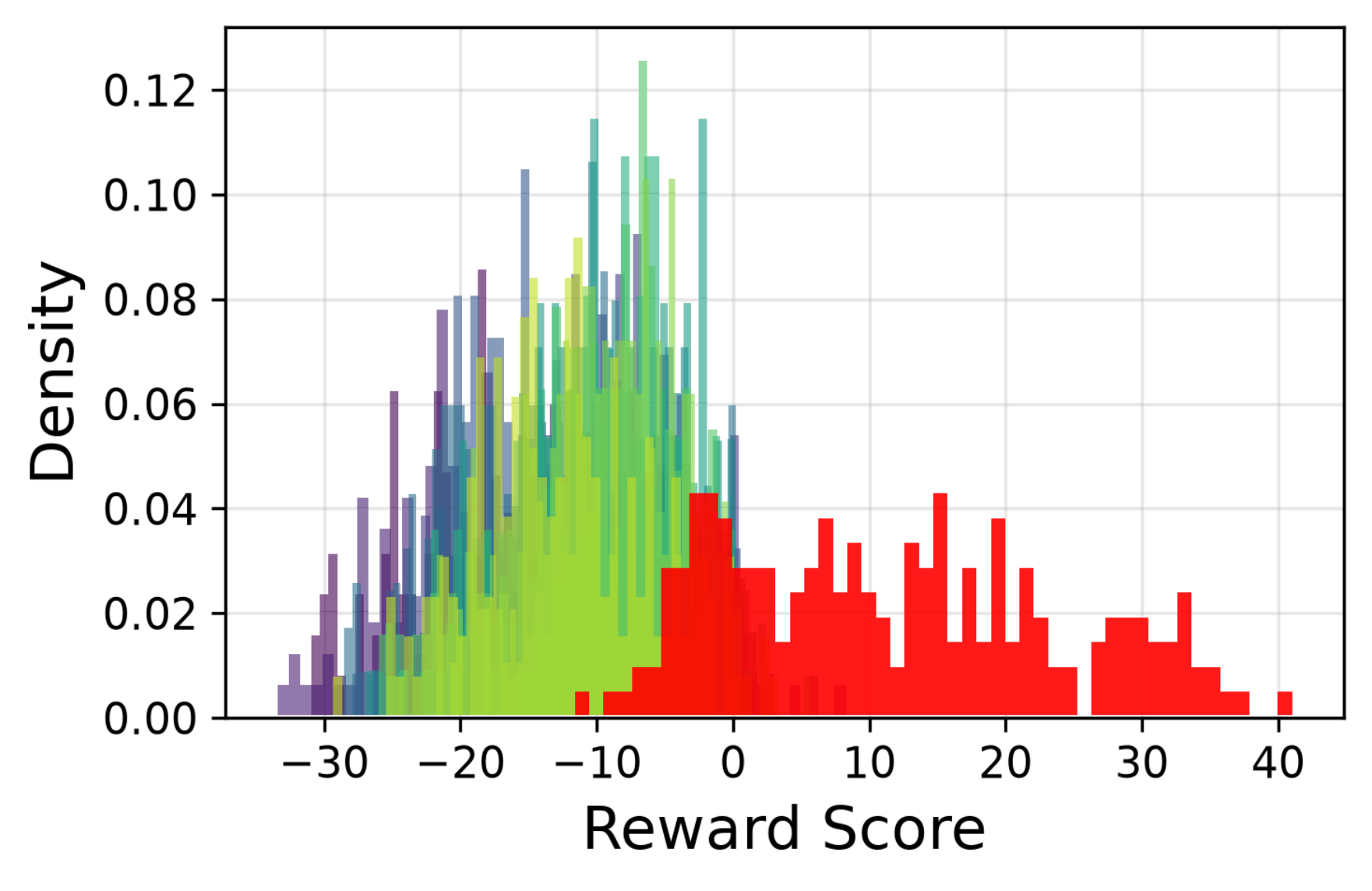}
    \\(c) After 2nd stage ($\alpha=0.75$)
  \end{minipage}
  
\vspace{-0.05in}
\caption{\textbf{Score distributions of the detector.} The detector is trained to identify Gemini-Pro. 
\textbf{\textcolor{red}{Red}} indicates the scores of the target model, and other colors indicate the score of non-target models.
(a) Initial detector before any fine-tuning. 
(b) After the first triplet training stage with interpolation factor $\alpha=0.5$, where the detector begins to assign higher scores to the target model; the score range spans approximately from $-20$ to $20$. 
(c) After iterative curriculum training with harder negatives ($\alpha=0.75$), resulting in clearer separation between target and non-target models; the score range further expands from about $-30$ to $40$.}

\vspace{-0.15in}
  \label{fig:gemini_distribution}
\end{figure*}

Table~\ref{tab:ablation_study} reports the effect of each component of \name{} on detector performance, averaged across three target LLMs (GPT-4o, Sonnet-4, and Gemini-Pro).
First, introducing triplet-based training improves Accuracy from 0.863 to 0.914 and AUROC from 0.882 to 0.931 under hard negative sampling, demonstrating the benefit of explicitly modeling relative preference orderings.
Second, adding adaptive curriculum learning further improves performance to 0.935 Accuracy and 0.951 AUROC.
Third, iterative curriculum training provides an additional improvement, with the full configuration achieving the best overall performance of 0.951 Accuracy and 0.958 AUROC.
Finally, the choice of negative sampling strategy also plays an important role.
Under the same triplet and adaptive training configuration, hard negative sampling achieves 0.935 Accuracy and 0.951 AUROC, outperforming both random sampling (0.928/0.920) and easy negative sampling (0.887/0.896).
Overall, these results show that triplet preference learning, adaptive curriculum learning, iterative training, and semantically challenging negatives each contribute to the final performance of \name{}.

\paragraph{Validation of interpolation-based difficulty.}

\begin{table}[t]
\centering
\small
\caption{
\textbf{Validation of the interpolation trajectory on GPT-4o}.
Target-response perplexity (PPL) decreases as $\alpha$ increases,
while generation-level embedding similarity (Emb. Sim.) increases.
}
\label{tab:interpolation_validation_gptonly}
\begin{tabular}{lcc}
\toprule
\textbf{Model} & PPL $\downarrow$ & Emb. Sim. $\uparrow$ \\
\midrule
Base (Phi-4)    & 2.114 & -- \\
$\alpha=0.25$   & 1.940 & 0.795 \\
$\alpha=0.50$   & 1.813 & 0.802 \\
$\alpha=0.75$   & 1.724 & 0.815 \\
Copy            & 1.668 & -- \\
\bottomrule
\end{tabular}
\vspace{-0.15in}
\end{table}

Our curriculum assumes that increasing the interpolation factor $\alpha$ produces progressively more target-like responses and therefore harder negatives.
To validate this ordering at the output level, we examine target-response perplexity (PPL) and generation-level embedding similarity along the
interpolation trajectory.

As shown in Table~\ref{tab:interpolation_validation_gptonly}, for GPT-4o, target PPL decreases monotonically as $\alpha$ increases toward the
copy-model endpoint, while embedding similarity to the target responses consistently increases.
The same monotonic trends are observed for Gemini-Pro and Sonnet-4; full results are in Appendix~\ref{sec:additional_interpolation_validation}.

These results provide output-level evidence that increasing $\alpha$ moves interpolated generations toward the target-mimicking endpoint.
Consequently, responses at $\alpha=0.5$ serve as relatively easier negatives, whereas those at $\alpha=0.75$ are more target-like and therefore harder to distinguish, supporting the curriculum ordering used in Section~\ref{sec:3.3}.


\paragraph{Score distributions.}

To gain deeper insight into how these performance improvements occur, we visualize the evolution of detector score distributions in Figure~\ref{fig:gemini_distribution}.
At the initial stage (a), the detector does not exhibit a clear preference toward the target model, resulting in substantial overlap between target and non-target responses.
In (b), after the first triplet training step with interpolation factor $\alpha=0.5$, the distribution of the target model begins to separate from the others, indicating that the detector has started to assign systematically higher scores to target responses.
Finally, in (c), iterative curriculum training with harder negatives ($\alpha=0.75$) produces a clearer margin between target and non-target distributions.
This progression demonstrates that \name{} gradually learns a target-specific scoring function.

\begin{table}[t]
\centering
\small
\caption{\textbf{Interpretability analysis via erasure-based attribution.}
We report the average reduction in discrimination margin after cumulatively removing the ten highest-ranked words on 20 high-margin GPT-4o examples.
Larger drops indicate that the removed words contribute more strongly to the detector's decision.}
\begin{tabular}{lc}
\toprule
\textbf{Deletion ranking} & \textbf{Margin drop} \\
\midrule
\name{}-ranked & \textbf{11.567} \\
TF--IDF-ranked & 6.452 \\
BoW-ranked & 6.394 \\
Random & 6.195 \\
Least-important & -0.690 \\
\bottomrule
\end{tabular}
\label{tab:interpretability_erasure}
\vspace{-0.15in}
\end{table}

\paragraph{Interpretability analysis.}
To examine which response cues drive \name{}'s decisions beyond those captured by shallow statistical baselines, we conduct an erasure-based local attribution analysis following \citet{Nguyen2018ComparingAA}.
We focus on GPT-4o test pairs for which \name{} correctly identifies the target response while TF--IDF incorrectly favors the non-target response, and select 20 high-margin cases for detailed analysis.

Following the discrimination margin defined in Eq.~\ref{eq:adaptive}, we use
$m_i = f_\theta(q^i, y_*^i) - f_\theta(q^i, y_o^i)$,
where $y_*^i$ is the GPT-4o response and $y_o^i$ denotes the paired non-target response at test time.
A positive $m_i$ therefore indicates that the detector assigns a higher target score to the actual GPT-4o response.
For each word, we measure its importance by the decrease in $m_i$ caused by deleting that word from the corresponding response.
We then cumulatively remove the ten highest-ranked words according to each attribution method and report the average resulting margin reduction.

As shown in Table~\ref{tab:interpretability_erasure}, removing words ranked most important by \name{} reduces the discrimination margin by 11.567 on average, substantially more than TF--IDF-ranked (6.452), BoW-ranked (6.394), or randomly selected words (6.195).
In contrast, removing the least-important words does not weaken the target decision.
These results indicate that \name{}'s predictions rely on decision-relevant response cues that are not fully characterized by shallow lexical statistics. A qualitative case study further examining these attribution cues is provided in Appendix~\ref{sec:qualitative_interpretability}.

\begin{table*}[t]
\vspace{-0.1in}
\centering
\small
\caption{\textbf{Required interactions (votes) to promote target models.}
We evaluate ranking manipulation across three target models with different initial ranks.
Each entry is reported as interactions (votes), where interactions indicate the total number of interactions and votes indicate the number of adversarial votes required to reach the target rank under different detectors.
Best interaction costs are highlighted in \textbf{bold}.}
\resizebox{\textwidth}{!}{%
\begin{tabular}{l cc cc cc}
\toprule
\multirow{3}{*}{\textbf{Method}}
& \multicolumn{2}{c}{\textbf{\texttt{o3-2025-04-16}}}
& \multicolumn{2}{c}{\textbf{\texttt{deepseek-v3-0324}}}
& \multicolumn{2}{c}{\textbf{\texttt{llama-4-scout-17b-16e-instruct}}} \\
& \multicolumn{2}{c}{\textbf{current rank: \#5}}
& \multicolumn{2}{c}{\textbf{current rank: \#15}}
& \multicolumn{2}{c}{\textbf{current rank: \#45}} \\
\cmidrule(lr){2-3} \cmidrule(lr){4-5} \cmidrule(lr){6-7}
& \textbf{Rank 1} & \textbf{Rank 2}
& \textbf{Rank 1} & \textbf{Rank 2}
& \textbf{Rank 1} & \textbf{Rank 2} \\
\midrule

Len-C
& 26400 (2188) & 25100 (2073)
& 76900 (6425) & 52400 (4389)
& 109100 (9142) & 67500 (5670) \\

Len-W
& 22200 (2137) & 6400 (588)
& 78400 (7645) & 44800 (4371)
& 99300 (9706) & 58900 (5741) \\

TF-IDF
& \textbf{21800} (2468) & 2500 (259)
& 72200 (8131) & 38300 (4307)
& 85300 (9686) & 51300 (5789) \\

BoW
& 22200 (2631) & 2300 (250)
& 69500 (8188) & 35800 (4254)
& 85200 (10082) & 49600 (5841) \\

Neural
& 22400 (2911) & 2300 (275)
& 62000 (8011) & 32700 (4262)
& 70600 (9124) & 45900 (5927) \\

Judge
& 22500 (2916) & 2300 (275)
& 62600 (8073) & 32800 (4269)
& 70600 (9100) & 45900 (5916) \\

\cellcolor{gray!40}\name{}
& \cellcolor{gray!40}22200 (3066)
& \cellcolor{gray!40}\textbf{2000} (243)
& \cellcolor{gray!40}\textbf{59600} (8137)
& \cellcolor{gray!40}\textbf{30900} (4257)
& \cellcolor{gray!40}\textbf{64700} (8871)
& \cellcolor{gray!40}\textbf{43300} (5912) \\

\bottomrule
\end{tabular}
}
\label{tab:rebuttal_simulation_baseline}
\end{table*}
\begin{table*}[t]
\vspace{-0.1in}
\centering
\small
\caption{
\textbf{Effect of competitor-aware voting with \name{}.}
We compare the target-only strategy and the enemy-list strategy for promoting the Rank~5 target model, \texttt{o3-2025-04-16}.
Each entry reports total interactions with adversarial votes in parentheses.
Interaction reduction is computed from total interactions.
}
\label{tab:competitor_aware_voting}
\resizebox{\textwidth}{!}{%
\begin{tabular}{lccccc}
\toprule
\textbf{Target model} &
\textbf{Current rank} &
\textbf{Target rank} &
\textbf{Target-only strategy} &
\textbf{Enemy-list strategy} &
\textbf{Interaction reduction} \\
\midrule
\multirow{4}{*}{\texttt{o3-2025-04-16}}
& \multirow{4}{*}{5}
& Rank~1 & 45{,}800 (1{,}618) & 22{,}200 (3{,}066) & 51.5\% \\
& 
& Rank~2 & 16{,}800 (564)     & 2{,}000 (243)      & 88.1\% \\
& 
& Rank~3 & 13{,}000 (442)     & 1{,}300 (158)      & 90.0\% \\
& 
& Rank~4 & 1{,}000 (37)       & 500 (53)           & 50.0\% \\
\bottomrule
\end{tabular}
}
\vspace{-0.1in}
\end{table*}

\section{Simulation of Ranking Manipulation}\label{sec:5}

To quantify the practical impact of model identification, we perform controlled simulations of ranking manipulation using real-world chatbot arena data.
Our goal is not to estimate the exact cost of manipulating a live platform, but to provide a controlled and ethically appropriate evaluation of how model identification affects the efficiency of strategic voting under fixed leaderboard dynamics.
Specifically, we adopt the setup by \citet{huang2025exploring}, which models adversarial voting efficiency under the BT model and Elo ranking dynamics. We use the \texttt{lmarena-ai/arena-human-preference-140k} dataset released in July 2025,\footnote{\url{https://news.lmarena.ai/opendata-july2025/}} containing approximately 140k pairwise battles. 
Rankings are reconstructed with BT model, and Elo scores are recalculated every 100 new votes. 
At each simulation step, the system draws two models uniformly at random from the candidate pool to form a synthetic battle. 
The attacker then receives two anonymized responses and chooses whether to cast a vote or abstain.

\paragraph{Attacker objectives and metrics.}
In the simulation, the attacker's objective is to shift a target model $M$ from its initial rank to a specified target rank (\textit{e.g.}, Rank 1). 
Then, the effectiveness of the attack is evaluated using two key metrics:
\begin{itemize}[leftmargin=3.5mm, itemsep=3pt, topsep=3pt, parsep=0.5pt]
    \item[$\circ$] \textbf{Total interactions:} The total number of queries submitted to the leaderboard. 
    This metric represents the overall resource cost, as the attacker must spend time generating responses even for battles where they eventually abstain.
    \item[$\circ$] \textbf{Adversarial votes:} The number of votes actually cast by the attacker. 
    This measures the direct influence exerted on the scoring system. 
    It is primarily determined by the BT ranking dynamics and remains relatively constant across different identification methods.
\end{itemize}

\paragraph{Simulation policy.}
For a target model $M$, we predefine an enemy list consisting of the top-ranked models in the initial leaderboard snapshot, specifically the models ranked 1--3.
This enemy list is kept fixed throughout the simulation, even as the target model's rank changes. 
Then, the attacker takes a following attack policy:
\begin{itemize}[leftmargin=3.5mm, itemsep=3pt, topsep=3pt, parsep=0.5pt]
    \item[$\circ$] \textit{When the target model is identified}, the attacker casts a vote that promotes the target.
    \item[$\circ$] \textit{When the target is not identified but the model on the enemy list is identified}, the attacker casts a vote against the identified enemy model.
    \item[$\circ$] \textit{Otherwise}, the attacker abstains.
\end{itemize}
Operationally, this strategy requires detectors only for the target model and the three fixed enemy models, avoiding the need to update the set of detectors as the leaderboard ranking changes.

\subsection{Simulation Results}
\label{sec:simulation_identification_methods}

Under the above policy, detector accuracy matters because a sampled battle is exploitable only when a present target or enemy-list model is correctly identified; identification errors make the attacker either spend interactions on misdirected votes or abstain from otherwise useful battles.
We therefore compare baseline detectors and \name{} as alternative identification modules for promoting the same target models.
The targets cover distinct model families and ranking regimes, with o3-2025-04-16, deepseek-v3-0324, and llama-4-scout-17b-16e-instruct initialized at ranks 5, 15, and 45, respectively.
For each target, we measure the total interactions and adversarial votes required to reach Rank~1 or Rank~2.
Table~\ref{tab:rebuttal_simulation_baseline} reports the required interactions and adversarial votes for each identification method under the same enemy-list voting policy.
Across most promotion objectives, \name{} achieves the lowest or competitive interaction cost, especially for lower-ranked targets where useful promotion opportunities are sparse.
Relative to the Length-Char detector, \name{} reduces interactions by 22.5\% for promoting deepseek-v3-0324 to Rank~1 and by 40.7\% for promoting llama-4-scout-17b-16e-instruct to Rank~1.
This demonstrates that stronger response identification directly lowers manipulation cost: weaker surface-statistic detectors miss more actionable battles and therefore require more Arena interactions, whereas \name{} converts sampled interactions into effective adversarial votes more efficiently.


\subsection{Effect of Competitor-Aware Voting}
\label{sec:simulations_with_interpol}

To assess whether identification enables manipulation beyond direct target promotion, we compare the enemy-list strategy with a target-only baseline.
The target-only baseline casts a vote only when the target model is identified and abstains otherwise, while the enemy-list strategy additionally downvotes identified higher-ranked competitors as defined above.
Table~\ref{tab:competitor_aware_voting} reports the comparison for promoting the Rank~5 target model, o3-2025-04-16.
For the most difficult objective, Rank~5$\rightarrow$1, the interaction cost decreases from 45{,}800 to 22{,}200, corresponding to a 51.5\% reduction.
Across the remaining objectives, the reduction ranges from 50.0\% to 90.0\%.
These results show that competitor-aware voting substantially lowers the operational cost of rank manipulation by exploiting not only direct target appearances but also battles involving higher-ranked competitors.

\section{Conclusion}
In this paper, we presented \name{}, a new model-driven approach to identify the source of LLM responses. 
\name{} leverages triplet-based training with adaptive and iterative curriculum learning, supported by synthetic negatives generated via model interpolation. 
Extensive experimental results demonstrate that \name{} consistently surpasses statistical baselines.
Our findings highlight a critical vulnerability in current anonymous voting-based leaderboards, showing that despite pragmatic defenses, substantial loopholes remain exploitable. 
By explicitly uncovering these weaknesses, our work emphasizes the necessity of moving beyond statistical safeguards and provides a foundation for designing robust and trustworthy evaluation infrastructures for LLMs.

\newpage

\section*{Limitations}

\name{} exposes residual response identifiability in anonymous LLM evaluation, but its current scope is centered on text-only, single-turn English interactions. 
This matches the main setting of Arena-style text leaderboards studied in this work, yet broader multilingual, multimodal, or multi-turn settings may introduce different attribution cues. 
Since \name{} only requires query--response pairs and target-specific training signals, extending the framework to these settings is a natural direction once suitable response corpora and detector backbones are available.

In addition, our manipulation analysis is based on reconstructed BT/Elo dynamics and controlled voting policies, rather than direct access to a deployed leaderboard system. 
The reported interaction counts should therefore be interpreted as relative simulation-based evidence of risk, not exact deployment-level attack costs. 
Incorporating platform-specific traffic patterns, aggregation rules, and defense mechanisms would further refine this estimate.

\section*{Broader Impact and Ethical Implications}

This work studies a vulnerability in anonymous, voting-based LLM evaluation. 
The intended impact of \name{} is defensive: it provides a way to test whether model responses remain anonymous under stronger identification methods than surface-level statistical baselines.
Such analysis can help leaderboard operators and evaluation researchers audit anonymity assumptions, detect manipulation risks, and design more robust aggregation and defense mechanisms.

At the same time, there is a risk that the capability studied in this paper can be misused.
A malicious user could apply response identification to infer which model produced an anonymous response and then cast selective votes to promote or suppress specific systems.
Our ranking-manipulation simulations demonstrate this risk under controlled conditions: stronger identification can reduce the number of interactions required for strategic voting.
Therefore, \name{} should not be used as a tool for manipulating public leaderboards.

All experiments in this paper are conducted offline using publicly available datasets, offline-collected model responses, reconstructed leaderboard dynamics, and simulated voting policies. 
We do not access private platform internals, user identities, or non-public metadata.
In practical deployments, response-identification systems should be used only as audit signals for evaluating platform robustness, not as standalone evidence for punitive action.
We encourage future work to use these findings to develop stronger anonymization, vote-auditing, and manipulation-resistant evaluation protocols.

\section*{Acknowledgments}
Jaehyung Kim is affiliated with the Department of Artificial Intelligence at Yonsei University.
This research was supported in part by the Institute for Information \& Communications Technology Planning \& Evaluation (IITP) grant funded by the Korea government (MSIT) (No. RS-2020-II201361, Artificial Intelligence Graduate School Program (Yonsei University); No. RS-2025-02215344, Development of AI Technology with Robust and Flexible Resilience Against Risk Factors).

\bibliography{custom}
\newpage
\appendix
\section{Experimental Details}

\subsection{Model Specification}
\label{tab:model specification}

The models used in our dataset are as follows:

\paragraph{Open-source models.}
\begin{itemize}[leftmargin=*, itemsep=0pt, topsep=2pt]
    \item Meta / Llama-3.2-3B-Instruct\footnote{\url{https://huggingface.co/meta-llama/Llama-3.2-3B-Instruct}}
    \item Meta / Llama-3.1-8B-Instruct\footnote{\url{https://huggingface.co/meta-llama/Llama-3.1-8B-Instruct}}
    \item Alibaba / Qwen2.5-3B-Instruct\footnote{\url{https://huggingface.co/Qwen/Qwen2.5-3B-Instruct}}
    \item Alibaba / Qwen3-8B\footnote{\url{https://huggingface.co/Qwen/Qwen3-8B}}
    \item Google / Gemma-2-2B-it\footnote{\url{https://huggingface.co/google/gemma-2-2b-it}}
    \item Google / Gemma-2-9B-it\footnote{\url{https://huggingface.co/google/gemma-2-9b-it}}
    \item Microsoft / Phi-4-mini-instruct\footnote{\url{https://huggingface.co/microsoft/Phi-4-mini-instruct}}
\end{itemize}

\paragraph{API models.}
\begin{itemize}[leftmargin=*, itemsep=0pt, topsep=2pt]
    \item Anthropic / Claude-4 Sonnet : \texttt{claude-sonnet-4-20250514}
    \item Anthropic / Claude-3.5 Sonnet : \texttt{claude-3-5-sonnet-20241022}
    \item Google / Gemini-Flash : \texttt{gemini-2.5-flash}
    \item Google / Gemini-Pro : \texttt{gemini-2.5-pro}
    \item OpenAI / GPT-4o : \texttt{gpt-4o-20240806}
    \item OpenAI / GPT-4o-mini : \texttt{gpt-4o-mini-20240718}
\end{itemize}

\paragraph{Embedding model.}
\begin{itemize}[leftmargin=*, itemsep=0pt, topsep=2pt]
    \item \url{sentence-transformers/all-MiniLM-L6-v2}
\end{itemize}

\subsection{Datasets and Metrics}
\label{tab:datasets and metrics}
For the experiments, we consider two different sources for input queries: (1) \textit{Alpaca dataset} \citep{alpaca}, an instruction-following corpus of approximately 52k examples generated by LLM widely used for fine-tuning purposes and (2) \textit{Arena human preference dataset} \citep{arena_data}, a large-scale collection of over 140k pairwise human-labeled preferences obtained from real-world interactions in an LM arena setting.
For the Alpaca dataset, we initiated a clustering process to read out 1,500 diverse samples and selected the final 1,400 for our query pool. 
For the Arena dataset, we filtered for single-turn, English-only conversations devoid of image inputs, randomly sampling 1,500 dialogues and retaining 1,400 after a cleaning process. 
We then gather responses to these queries from a diverse set of 12 LLMs, which can be grouped into six families, each containing two variants: Llama-3.2-3B/Llama-3.1-8B~\citep{grattafiori2024llama}, Qwen2.5-3B/Qwen3-8B~\citep{qwen2_5, qwen3}, Gemma-2-2B/Gemma-2-9B~\citep{team2024gemma}, Claude-Sonnet-4/Claude-Sonnet-3.5~\citep{claude4_system_card, claude35_sonnet_announcement}, Gemini-Flash/Gemini-Pro~\citep{comanici2025gemini25pushingfrontier}, and GPT-4o-mini/GPT-4o~\citep{gpt4o_report}.

The resulting dataset, comprising query-response pairs, is pre-split into training
(1,000), validation (200), and test (200) sets using a fixed random seed (seed=42) to ensure reproducibility and prevent data leakage. 
For training, we employ a sentence encoder to select the most semantically similar response from other models, which serves as the negative sample in triplet construction. 
Meanwhile, for validation and testing, we evaluate each query against all non-target models, resulting in 200 queries $\times$ 11 models = 2,200 scored pairs per split. 
Our main experimental setup focuses on three primary target LLMs: GPT-4o, Gemini-Pro, and Sonnet-4. 
For evaluation, we report both Accuracy and AUROC on the test sets by using validation sets for model selection. 
These metrics are computed and averaged across the 11 non-target LLMs to assess the generalization performance of our detector.

\subsection{Baseline Details}
\label{app:baseline_details}

\paragraph{Statistical baselines.}
We evaluate four feature-based baselines trained with $\ell_2$-regularized logistic regression (SAGA solver, $C{=}1.0$), selecting the best seed by validation AUROC over five seeds (100, 200, 300, 400, 500).
\begin{itemize}[leftmargin=*, itemsep=0pt, topsep=2pt]
    \item \textbf{Length–Word}: number of whitespace-separated tokens in a response; features are standardized.
    \item \textbf{Length–Char}: number of characters in a response; features are standardized.
    \item \textbf{BoW}: bag-of-words counts with n-grams up to 2 and \texttt{min\_df}=2.
    \item \textbf{TF–IDF}: TF–IDF features with the same n-gram setting (1–2) and \texttt{min\_df}=2.
\end{itemize}
For each baseline we train until convergence with a sufficiently large iteration cap (up to 2{,}000).

\paragraph{Neural-based multi-class classification baseline.}

We implemented a neural network-based baseline in which a pre-trained transformer model, \texttt{DeBERTa-v3-base}, is fine-tuned to classify the source of each generated response among all candidate LLMs following the source-attribution formulation of \citet{tay2020reverse}. 
Given a pool of $N$ language models, the classifier is trained to predict which model produced a given answer. The training dataset comprises query-response pairs where each query is answered by all $N$ models, yielding $N$ labeled samples per query. We employ standard supervised learning with cross-entropy loss and select the best checkpoint based on validation accuracy. For evaluation, we apply the model to pairs of candidate responses (e.g., a response from the target model and another from a competitor model) and compare the predicted logits. The response receiving a higher logit for the target class is considered to be from the target model. To reduce position bias, each test pair is evaluated in both forward and reversed order, and only consistent predictions are counted as correct.

\paragraph{LLM-as-a-Judge baseline with few-shot in-context learning.}
As another strong baseline, we utilize an open-source LLM as an automated judge.
This baseline leverages the in-context learning ability of LLM by providing several few-shot exemplars sampled from the training data. 
Each prompt consists of an instruction and two candidate responses (the target and a distractor), along with the ground-truth label indicating which response was produced by the target model. 
For each test case, we query the judge model twice—once in each order (target vs. other and other vs. target)—while keeping the few-shot context fixed for all queries. 
The judge is instructed to identify the response most likely to have been generated by the target model. 
Only cases where the judge correctly identifies the target regardless of position are scored as correct; ambiguous or contradictory cases receive partial credit. 
We implement this baseline using Gemma-2-27B-it~\cite{team2024gemma}. 
To ensure computational efficiency, we utilize vLLM's prefix caching feature, as the lengthy few-shot prompt remains identical across all inference requests.

\begin{table*}[!t]
\centering
\small
\caption{\textbf{Model-wise Accuracy on Arena human preference dataset.}
Abbreviations: C3.5 = Sonnet-3.5, C4 = Sonnet-4, G-F = Gemini-Flash, G-P = Gemini-Pro, 
G2-2B = Gemma-2-2B-it, G2-9B = Gemma-2-9B-it, GPT-M = GPT-4o-mini, GPT-4 = GPT-4o, 
L3.1-8B = Llama-3.1-8B-Instruct, L3.2-3B = Llama-3.2-3B-Instruct, 
Q2.5-3B = Qwen2.5-3B-Instruct, Q3-8B = Qwen3-8B. 
Cells highlighted in \textcolor{red}{red} indicate \emph{family models} of the target high model 
(\textit{e.g.}, GPT-4o vs GPT-4o-mini). The best scores are highlighted in \textbf{bold}.}
\resizebox{\textwidth}{!}{%
\begin{tabular}{l l c c c c c c c c c c c c}
\toprule
\textbf{Target} & \textbf{Method} & C3.5 & C4 & G-F & G-P & G2-2B & G2-9B & GPT-M & GPT-4 & L3.1-8B & L3.2-3B & Q2.5-3B & Q3-8B \\
\midrule

GPT-4o  & Length-Word & 0.385 & 0.415 & 0.665 & 0.708 & 0.613 & 0.488 &\cellcolor{red!20} 0.515 & --    & 0.588 & 0.565 & 0.543 & 0.635 \\
        & Length-Char & 0.395 & 0.428 & 0.675 & 0.713 & 0.620 & 0.508 &\cellcolor{red!20} 0.513 & --    & 0.588 & 0.575 & 0.555 & 0.633 \\
        & TF-IDF   & 0.788 & 0.778 & 0.700 & 0.725 & 0.690 & 0.695 &\cellcolor{red!20} 0.645 & --    & 0.665 & 0.663 & 0.585 & 0.733 \\
        & BoW      & 0.750 & 0.775 & 0.708 & 0.733 & 0.715 & 0.693 &\cellcolor{red!20} 0.690 & --    & 0.698 & 0.690 & 0.605 & 0.748 \\
        & Neural-Based      & 0.865 & 0.965 & \textbf{0.970} & \textbf{0.980} & \textbf{0.985} & 0.985 &\cellcolor{red!20} 0.590 & --    & \textbf{0.960} & \textbf{0.965} & 0.660 & \textbf{0.955} \\
        & LLM-judge      & 0.710 & 0.730 & 0.750 & 0.825 & 0.830 & 0.750 &\cellcolor{red!20} \textbf{0.870} & --    & 0.890 & 0.915 & \textbf{0.915} & 0.855 \\
        & \cellcolor{gray!20}\name{}
        & \cellcolor{gray!20}\textbf{0.875}
        & \cellcolor{gray!20}\textbf{0.970}
        & \cellcolor{gray!20}0.965
        & \cellcolor{gray!20}0.970
        & \cellcolor{gray!20}0.975
        & \cellcolor{gray!20}\textbf{0.990}
        & \cellcolor{red!20}0.740
        & \cellcolor{gray!20}--
        & \cellcolor{gray!20}0.900
        & \cellcolor{gray!20}0.920
        & \cellcolor{gray!20}0.740
        & \cellcolor{gray!20}0.945 \\
\midrule

Gemini-Pro & Length-Word & 0.835 & 0.793 &\cellcolor{red!20} 0.585 & --    & 0.830 & 0.858 & 0.848 & 0.813 & 0.785 & 0.805 & 0.825 & 0.733 \\
           & Length-Char & 0.805 & 0.770 &\cellcolor{red!20} 0.570 & --    & 0.795 & 0.833 & 0.825 & 0.788 & 0.768 & 0.778 & 0.795 & 0.723 \\
           & TF-IDF   & 0.905 & 0.883 &\cellcolor{red!20} 0.728 & --    & 0.833 & 0.870 & 0.860 & 0.845 & 0.838 & 0.830 & 0.853 & 0.840 \\
           & BoW      & 0.890 & 0.885 & \cellcolor{red!20} 0.818 & --    & 0.903 & 0.913 & 0.895 & 0.898 & 0.863 & 0.900 & 0.905 & 0.885 \\
           & Neural-Based      & 0.955 & 0.955 & \cellcolor{red!20} 0.750 & --    & 0.905 & 0.900 & 0.915 & 0.935 & 0.920 & 0.940 & 0.915 & 0.895 \\
           & LLM-judge      & 0.960 & 0.955 & \cellcolor{red!20} 0.930 & --    & 0.960 & 0.940 & 0.955 & 0.955 & 0.970 & 0.970 & 0.975 & 0.970 \\
           & \cellcolor{gray!20}\name{}
           & \cellcolor{gray!20}\textbf{1.000}
           & \cellcolor{gray!20}\textbf{1.000}
           & \cellcolor{red!20}\textbf{1.000}
           & \cellcolor{gray!20}--
           & \cellcolor{gray!20}\textbf{1.000}
           & \cellcolor{gray!20}\textbf{1.000}
           & \cellcolor{gray!20}\textbf{1.000}
           & \cellcolor{gray!20}\textbf{1.000}
           & \cellcolor{gray!20}\textbf{1.000}
           & \cellcolor{gray!20}\textbf{1.000}
           & \cellcolor{gray!20}\textbf{1.000}
           & \cellcolor{gray!20}\textbf{1.000} \\
\midrule

Sonnet 4   & Length-Word &\cellcolor{red!20} 0.465 & --    & 0.753 & 0.803 & 0.683 & 0.553 & 0.560 & 0.593 & 0.655 & 0.643 & 0.600 & 0.715 \\
           & Length-Char &\cellcolor{red!20} 0.538 & --    & 0.255 & 0.205 & 0.350 & 0.468 & 0.475 & 0.428 & 0.370 & 0.378 & 0.408 & 0.305 \\
           & TF-IDF   &\cellcolor{red!20} 0.583 & --    & 0.818 & 0.845 & 0.865 & 0.840 & 0.850 & 0.825 & 0.820 & 0.830 & 0.823 & 0.818 \\
           & BoW      &\cellcolor{red!20} 0.635 & --    & 0.825 & 0.868 & 0.890 & 0.870 & 0.880 & 0.898 & 0.893 & 0.875 & 0.873 & 0.853 \\
           & Neural-Based       &\cellcolor{red!20} 0.830 & --    & 0.910 & 0.935 & 0.930 & 0.930 & 0.850 & 0.860 & 0.865 & 0.875 & 0.860 & 0.865 \\
           & LLM-judge      &\cellcolor{red!20} 0.860 & --    & 0.835 & 0.905 & 0.905 & 0.845 & \textbf{0.925} & \textbf{0.925} & 0.935 & 0.950 & \textbf{0.925} & 0.910 \\
           & \cellcolor{gray!20}\name{}
           & \cellcolor{red!20}\textbf{0.915}
           & \cellcolor{gray!20}--
           & \cellcolor{gray!20}\textbf{0.970}
           & \cellcolor{gray!20}\textbf{0.955}
           & \cellcolor{gray!20}\textbf{0.975}
           & \cellcolor{gray!20}\textbf{0.975}
           & \cellcolor{gray!20}0.920
           & \cellcolor{gray!20}0.920
           & \cellcolor{gray!20}\textbf{0.960}
           & \cellcolor{gray!20}\textbf{0.955}
           & \cellcolor{gray!20}0.920
           & \cellcolor{gray!20}\textbf{0.940} \\
\bottomrule
\end{tabular}
}

\label{tab:modelwise_human_accuracy}
\end{table*}
\begin{table*}[!t]
\centering
\small
\caption{\textbf{Model-wise AUROC on human preference dataset.}
Abbreviations: C3.5 = Sonnet-3.5, C4 = Sonnet-4, G-F = Gemini-Flash, G-P = Gemini-Pro, 
G2-2B = Gemma-2-2B-it, G2-9B = Gemma-2-9B-it, GPT-M = GPT-4o-mini, GPT-4 = GPT-4o, 
L3.1-8B = Llama-3.1-8B-Instruct, L3.2-3B = Llama-3.2-3B-Instruct, 
Q2.5-3B = Qwen2.5-3B-Instruct, Q3-8B = Qwen3-8B. 
Cells highlighted in \textcolor{red}{red} indicate \emph{family models} of the target high model 
(e.g., GPT-4o vs GPT-4o-mini). The best scores are highlighted in \textbf{bold}.}
\resizebox{\textwidth}{!}{%
\begin{tabular}{l l c c c c c c c c c c c c}
\toprule
Target & Method & C3.5 & C4 & G-F & G-P & G2-2B & G2-9B & GPT-M & GPT-4 & L3.1-8B & L3.2-3B & Q2.5-3B & Q3-8B \\
\midrule

GPT-4o  & Length-Word & 0.344 & 0.462 & 0.800 & 0.879 & 0.652 & 0.503 & \cellcolor{red!20}0.535 & -- & 0.642 & 0.610 & 0.571 & 0.721 \\
        & Length-Char & 0.351 & 0.479 & 0.795 & 0.867 & 0.650 & 0.513 & \cellcolor{red!20}0.534 & -- & 0.632 & 0.602 & 0.576 & 0.706 \\
        & TF-IDF & 0.866 & 0.883 & 0.780 & 0.808 & 0.749 & 0.755 & \cellcolor{red!20}0.713 & -- & 0.716 & 0.719 & 0.616 & 0.817 \\
        & BoW & 0.827 & 0.855 & 0.775 & 0.819 & 0.756 & 0.752 & \cellcolor{red!20}\textbf{0.776} & -- & 0.733 & 0.745 & 0.633 & 0.823 \\
        & Neural-Based & 0.890 & 0.956 & \textbf{0.966} & \textbf{0.984} & 0.981 & 0.975 & \cellcolor{red!20}0.546 & -- & \textbf{0.952} & \textbf{0.965} & 0.610 & 0.952 \\
        & \cellcolor{gray!20}\name{}
        & \cellcolor{gray!20}\textbf{0.902}
        & \cellcolor{gray!20}\textbf{0.980}
        & \cellcolor{gray!20}0.961
        & \cellcolor{gray!20}0.978
        & \cellcolor{gray!20}\textbf{0.982}
        & \cellcolor{gray!20}\textbf{0.977}
        & \cellcolor{red!20}0.762
        & \cellcolor{gray!20}--
        & \cellcolor{gray!20}0.916
        & \cellcolor{gray!20}0.934
        & \cellcolor{gray!20}\textbf{0.762}
        & \cellcolor{gray!20}\textbf{0.969} \\
\midrule

Gemini-Pro & Length-Word & 0.910 & 0.845 & \cellcolor{red!20}0.565 & -- & 0.844 & 0.885 & 0.879 & 0.860 & 0.834 & 0.844 & 0.858 & 0.806 \\
           & Length-Char & 0.896 & 0.817 & \cellcolor{red!20}0.542 & -- & 0.826 & 0.872 & 0.867 & 0.844 & 0.823 & 0.836 & 0.843 & 0.795 \\
           & TF-IDF & 0.965 & 0.957 & \cellcolor{red!20}0.848 & -- & 0.919 & 0.940 & 0.930 & 0.932 & 0.917 & 0.923 & 0.935 & 0.919 \\
           & BoW & 0.952 & 0.937 & \cellcolor{red!20}0.905 & -- & 0.955 & 0.959 & 0.957 & 0.958 & 0.944 & 0.955 & 0.957 & 0.948 \\
           & Neural-Based & 0.946 & 0.945 & \cellcolor{red!20}0.891 & -- & 0.941 & 0.940 & 0.947 & 0.947 & 0.947 & 0.948 & 0.947 & 0.946 \\
           & \cellcolor{gray!20}\name{}
           & \cellcolor{gray!20}\textbf{0.983}
           & \cellcolor{gray!20}\textbf{0.987}
           & \cellcolor{red!20}\textbf{0.972}
           & \cellcolor{gray!20}--
           & \cellcolor{gray!20}\textbf{0.984}
           & \cellcolor{gray!20}\textbf{0.985}
           & \cellcolor{gray!20}\textbf{0.975}
           & \cellcolor{gray!20}\textbf{0.975}
           & \cellcolor{gray!20}\textbf{0.976}
           & \cellcolor{gray!20}\textbf{0.977}
           & \cellcolor{gray!20}\textbf{0.973}
           & \cellcolor{gray!20}\textbf{0.984} \\
\midrule

Sonnet 4   & Length-Word & \cellcolor{red!20}0.342 & -- & 0.783 & 0.845 & 0.659 & 0.553 & 0.539 & 0.550 & 0.647 & 0.623 & 0.587 & 0.699 \\
           & Length-Char & \cellcolor{red!20}0.662 & -- & 0.235 & 0.183 & 0.357 & 0.463 & 0.479 & 0.467 & 0.374 & 0.402 & 0.426 & 0.324 \\
           & TF-IDF & \cellcolor{red!20}0.660 & -- & 0.908 & 0.947 & 0.950 & 0.931 & 0.937 & 0.934 & 0.913 & 0.910 & 0.925 & 0.906 \\
           & BoW & \cellcolor{red!20}0.756 & -- & 0.891 & 0.942 & 0.969 & 0.952 & 0.962 & 0.962 & 0.964 & 0.935 & 0.948 & 0.938 \\
           & Neural-Based & \cellcolor{red!20}0.858 & -- & 0.904 & 0.914 & 0.907 & 0.909 & 0.916 & 0.918 & 0.918 & 0.919 & 0.919 & 0.908 \\
           & \cellcolor{gray!20}\name{}
           & \cellcolor{red!20}\textbf{0.961}
           & \cellcolor{gray!20}--
           & \cellcolor{gray!20}\textbf{0.975}
           & \cellcolor{gray!20}\textbf{0.979}
           & \cellcolor{gray!20}\textbf{0.991}
           & \cellcolor{gray!20}\textbf{0.989}
           & \cellcolor{gray!20}\textbf{0.969}
           & \cellcolor{gray!20}\textbf{0.964}
           & \cellcolor{gray!20}\textbf{0.980}
           & \cellcolor{gray!20}\textbf{0.984}
           & \cellcolor{gray!20}\textbf{0.963}
           & \cellcolor{gray!20}\textbf{0.976} \\
\bottomrule
\end{tabular}
}

\label{tab:modelwise_human_auroc}
\end{table*}

\subsection{Implementation Details}
\label{Training and implementation details}


To construct the query–response corpus required for detector training, we generate responses for 1,400 queries from a diverse pool of 12 LLMs (12\,$\times$\,1{,}400\,=\,16{,}800 generations) using a unified decoding protocol. 
Local generations are produced with a high\mbox{-}throughput vLLM engine configured with sampling temperature $0.7$, top\mbox{-}$p=0.9$, a maximum of $4096$ generated tokens, and stop sequences $\{\texttt{<|eot\_id|>}, \texttt{<|end\_of\_text|>}\}$. 
For stability in cross\mbox{-}model comparisons, API\mbox{-}based generations are obtained with temperature fixed to $0$ (greedy decoding). 


We adopt GRM-Llama-3.2-3B%
\footnote{\href{https://huggingface.co/Ray2333/GRM-Llama3.2-3B-rewardmodel-ft}
{Hugging Face model page}},
which is a Llama-3.2-3B-Instruct model fine-tuned for reward modeling and serves as our primary detector backbone. Its training is performed with the AdamW optimizer ($\beta_1=0.9, \beta_2=0.999, \epsilon=1\times10^{-8}$) and a weight decay of 0.01. We employ a cosine learning rate schedule with 100 warmup steps.
Training follows a two-stage curriculum, with each stage running for 5 epochs.
Stage 1 trains on easier triplets constructed with interpolated negatives at $\alpha_1=0.5$, using an initial learning rate of $1\times10^{-5}$. 
Stage 2 continues from the best Stage 1 checkpoint, training on harder triplets at $\alpha_2=0.75$ with a halved learning rate of $5\times10^{-6}$.
We use a batch size of 2 with gradient accumulation of 4 (effective batch size 8) and FP16 mixed precision, leveraging DeepSpeed ZeRO-2 for memory efficiency. 
A weighted triplet loss is applied with coefficients $(\lambda_1, \lambda_2, \lambda_3) = (0.3, 0.3, 1.0)$. The final model checkpoint is selected based on the highest AUROC on the validation set. 

For the copymodel, we fine-tune Phi-4-mini-instruct on the same pre-split datasets. 
The model is trained for 3 epochs with a batch size of 4 and a learning rate of $2\times10^{-5}$. 
The maximum sequence length is set to 512, and the best-performing checkpoint is selected based on the lowest validation loss. 
For the main development, we used a single NVIDIA RTX 6000 Ada 96GB GPU.


\section{Additional Results}
\label{app:Modelwise Results}

\subsection{Model-wise Breakdown Accuracy}
\label{app:Modelwise breakdown Accuracy}

We further examine robustness under within-family model comparisons in Table~\ref{tab:modelwise_human_accuracy}.
Across all target LLMs, most existing methods exhibit noticeable accuracy drop when distinguishing closely related family variants.
This degradation is particularly pronounced for feature-based statistical baselines, which rely on superficial cues and often fail to separate models sharing similar training data, architectures, or distillation pipelines. 
In contrast, \name{} consistently maintains high identification accuracy in these challenging within-family settings.
For example, when the target is GPT-4o or Sonnet-4, statistical baselines such as TF-IDF and BoW show substantial drops in accuracy against their respective family models, whereas \name{} preserves strong and stable performance.

\subsection{Model-wise Breakdown AUROC}
\label{app:Modelwise breakdown Auroc}

In Table~\ref{tab:modelwise_human_auroc}, we report model-wise AUROC results on the Arena Human preference dataset. 
Beyond simple classification accuracy, AUROC provides a richer and threshold-independent view of separability, capturing how consistently the detector assigns higher scores to target responses compared to non-target ones. 
The consistently high AUROC values achieved by \name{} across all non-target models confirm that our method offers not only strong accuracy but also robust and reliable discrimination, even in challenging settings such as family models (\textit{e.g.}, GPT-4o vs. GPT-4o-mini, Sonnet-4 vs. Sonnet-3.5).

\subsection{Close-Elo Competitor Evaluation}
\label{app:close_elo_breakdown}

To further examine whether \name{} relies on broad capability gaps between models, we construct a close-Elo competitor pool from the official LM Arena text leaderboard.
The leaderboard snapshot was captured on \texttt{2025-11-21}. In addition to our three target models, GPT-4o, Gemini-Pro, and Sonnet-4, we include five high-ranked competitors with similar Elo ratings: GLM-4.6 (Rank 11; Elo 1428), Qwen3-Max-2025-09-23 (Rank 13; Elo 1423), Grok-4-Fast (Rank 19; Elo 1420), Kimi-K2-0905-preview (Rank 21; Elo 1416), and DeepSeek-V3.1 (Rank 22; Elo 1416).

\begin{table}[!t]
\centering
\small

\caption{
\textbf{Target-present identification results in the close-Elo Arena setting.}
We rerun representative baselines among models with highly similar Arena Elo scores, focusing on direct competitors rather than wide capability gaps.
We report average Accuracy (Acc.) and AUROC across seven close-Elo non-target LLMs.
Best scores are highlighted in \textbf{bold}.
Method abbreviations: Len-W = Length--Word, Len-C = Length--Char, TF-IDF = TF--IDF, BoW = Bag-of-Words, and Neural = Neural-based classification.
}

\label{tab:close_elo_baseline}
\resizebox{\columnwidth}{!}{%
\begin{tabular}{@{}llcc@{}}
\toprule
\multirow{2}{*}{\textbf{Target}} 
& \multirow{2}{*}{\textbf{Method}} 
& \multicolumn{2}{c}{\textbf{Close-Elo Arena}} \\
\cmidrule(lr){3-4}
& & Acc. & AUROC \\
\midrule

\multirow{6}{*}{GPT-4o}
& Len-W   & 0.491 & 0.517 \\
& Len-C   & 0.495 & 0.518 \\
& TF-IDF  & 0.755 & 0.848 \\
& BoW     & 0.773 & 0.869 \\
& Neural  & 0.766 & 0.852 \\
& \cellcolor{gray!20}\name{} 
& \cellcolor{gray!20}\textbf{0.949} 
& \cellcolor{gray!20}\textbf{0.932} \\
\midrule

\multirow{6}{*}{Gemini-Pro}
& Len-W   & 0.632 & 0.671 \\
& Len-C   & 0.627 & 0.663 \\
& TF-IDF  & 0.713 & 0.782 \\
& BoW     & 0.669 & 0.703 \\
& Neural  & \textbf{0.851} & \textbf{0.811} \\
& \cellcolor{gray!20}\name{} 
& \cellcolor{gray!20}0.806 
& \cellcolor{gray!20}0.775 \\
\midrule

\multirow{6}{*}{Sonnet-4}
& Len-W   & 0.536 & 0.493 \\
& Len-C   & 0.523 & 0.486 \\
& TF-IDF  & 0.705 & 0.767 \\
& BoW     & 0.762 & 0.872 \\
& Neural  & 0.882 & 0.931 \\
& \cellcolor{gray!20}\name{} 
& \cellcolor{gray!20}\textbf{0.989} 
& \cellcolor{gray!20}\textbf{0.975} \\
\bottomrule
\end{tabular}
}
\end{table}

As shown in Table~\ref{tab:close_elo_baseline}, this close-Elo setting is substantially harder than the standard broad-pool evaluation. \name{} achieves the strongest average Accuracy for GPT-4o and Sonnet-4, reaching 0.949 and 0.989 respectively, while Gemini-Pro becomes more challenging and the Neural baseline is slightly stronger. Nevertheless, \name{} remains competitive for Gemini-Pro and still outperforms feature-based baselines such as Length, BoW, and TF-IDF. This indicates that restricting the evaluation to high-ranked, similarly rated competitors does not remove the attribution signal learned by \name{}.

\begin{table*}[!t]
\vspace{-0.1in}
\centering
\small
\caption{\textbf{Model-wise Accuracy in the close-Elo pool setting.}
We evaluate \name{} among models with highly similar Arena Elo scores, focusing on direct competitors rather than wide capability gaps.
The Avg. column reports the mean across seven close-Elo non-target models.}
\resizebox{\textwidth}{!}{%
\begin{tabular}{l c c c c c c c c c}
\toprule
\textbf{Target} 
& \textbf{GPT-4o} 
& \textbf{Gemini-Pro} 
& \textbf{Sonnet-4} 
& \textbf{Grok-4-Fast} 
& \textbf{DeepSeek-V3.1} 
& \textbf{GLM-4.6} 
& \textbf{Kimi-K2} 
& \textbf{Qwen3-Max}
& \textbf{Avg.} \\
\midrule
GPT-4o      
& --    
& 0.985 
& 0.975 
& 0.935 
& 0.950 
& 0.975 
& 0.950 
& 0.875 
& 0.949 \\

Gemini-Pro  
& 0.825 
& --    
& 0.925 
& 0.880 
& 0.705 
& 0.580 
& 0.860 
& 0.870 
& 0.806 \\

Sonnet-4    
& 0.975 
& 1.000 
& --    
& 0.985 
& 0.995 
& 0.995 
& 0.990 
& 0.980 
& 0.989 \\
\bottomrule
\end{tabular}
}
\label{tab:close_elo_modelwise_accuracy}
\end{table*}
\begin{table*}[!t]
\vspace{-0.1in}
\centering
\small
\caption{\textbf{Model-wise AUROC in the close-Elo pool setting.}
We evaluate \name{} among models with highly similar Arena Elo scores, focusing on direct competitors rather than wide capability gaps.
The Avg. column reports the mean across seven close-Elo non-target models.}
\resizebox{\textwidth}{!}{%
\begin{tabular}{l c c c c c c c c c}
\toprule
\textbf{Target} 
& \textbf{GPT-4o} 
& \textbf{Gemini-Pro} 
& \textbf{Sonnet-4} 
& \textbf{Grok-4-Fast} 
& \textbf{DeepSeek-V3.1} 
& \textbf{GLM-4.6} 
& \textbf{Kimi-K2} 
& \textbf{Qwen3-Max}
& \textbf{Avg.} \\
\midrule
GPT-4o      
& --    
& 0.976 
& 0.959 
& 0.902 
& 0.936 
& 0.960 
& 0.930 
& 0.858 
& 0.932 \\

Gemini-Pro  
& 0.781 
& --    
& 0.872 
& 0.853 
& 0.684 
& 0.554 
& 0.827 
& 0.854 
& 0.775 \\

Sonnet-4    
& 0.967 
& 0.985 
& --    
& 0.970 
& 0.973 
& 0.985 
& 0.973 
& 0.969 
& 0.975 \\
\bottomrule
\end{tabular}
}
\label{tab:close_elo_modelwise_auroc}
\vspace{-0.1in}
\end{table*}

Tables~\ref{tab:close_elo_modelwise_accuracy} and~\ref{tab:close_elo_modelwise_auroc} provide the full pairwise model-level breakdowns for \name{}.
For GPT-4o and Sonnet-4, \name{} maintains consistently high pairwise performance across nearly all close-Elo opponents.
Gemini-Pro shows lower performance against several nearby competitors, especially glm-4.6 and DeepSeek-v3.1, suggesting that this target constitutes the hardest case in the close-Elo pool.
Overall, these results support that \name{} is not driven only by broad model-quality gaps, but captures target-specific response signatures that persist among closely rated Arena models.

\section{Detailed Analyses}\label{appendix:c}

\subsection{Additional Score Distributions}
\label{sec:additional-score-distributions}

\begin{figure*}[t]
    \vspace{-0.1in}
  \centering
  \begin{minipage}[t]{0.32\linewidth}
    \centering
    \includegraphics[width=\linewidth]{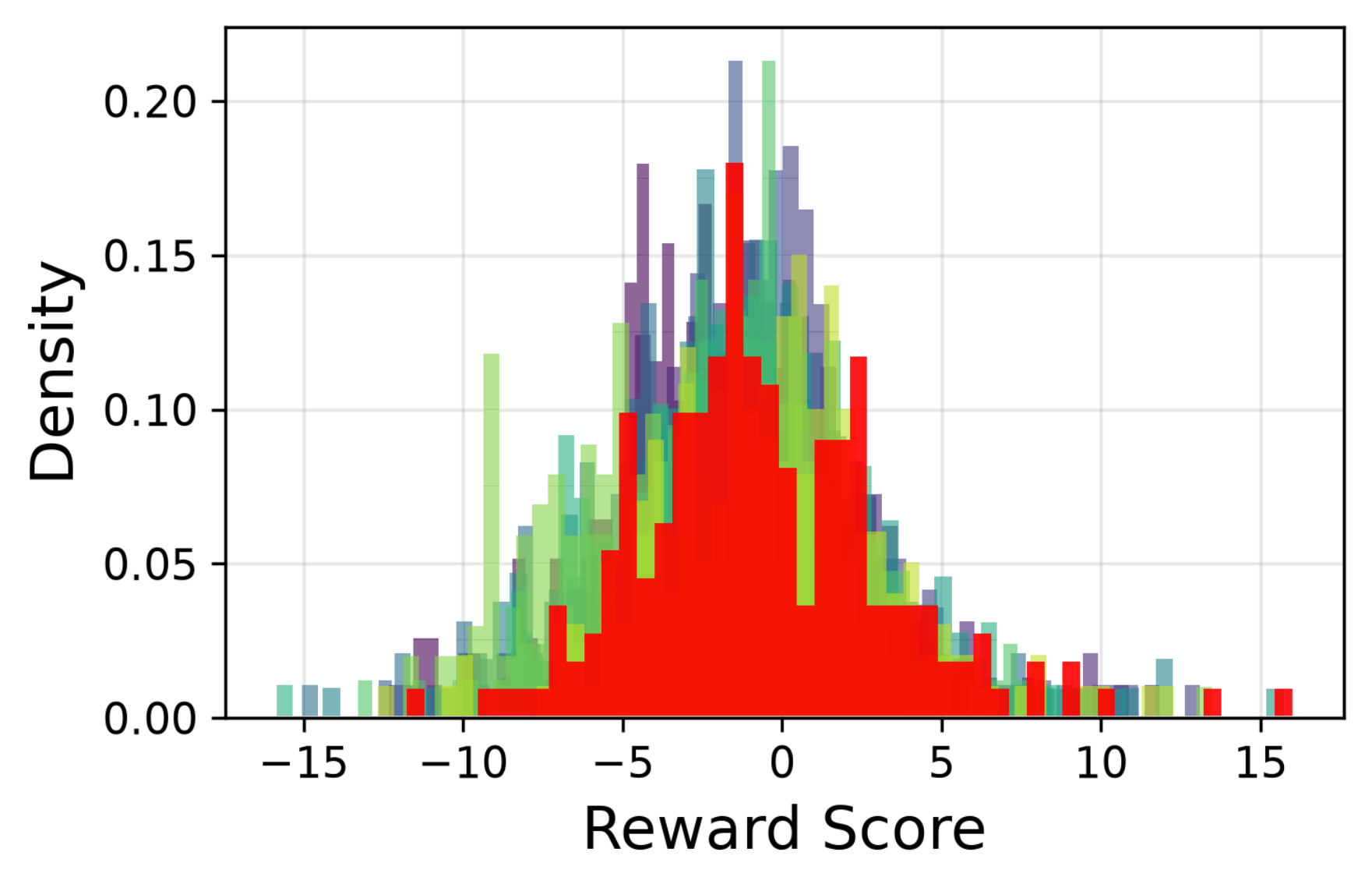}
    \\(a) Initial detector
  \end{minipage}
  \hfill
  \begin{minipage}[t]{0.32\linewidth}
    \centering
    \includegraphics[width=\linewidth]{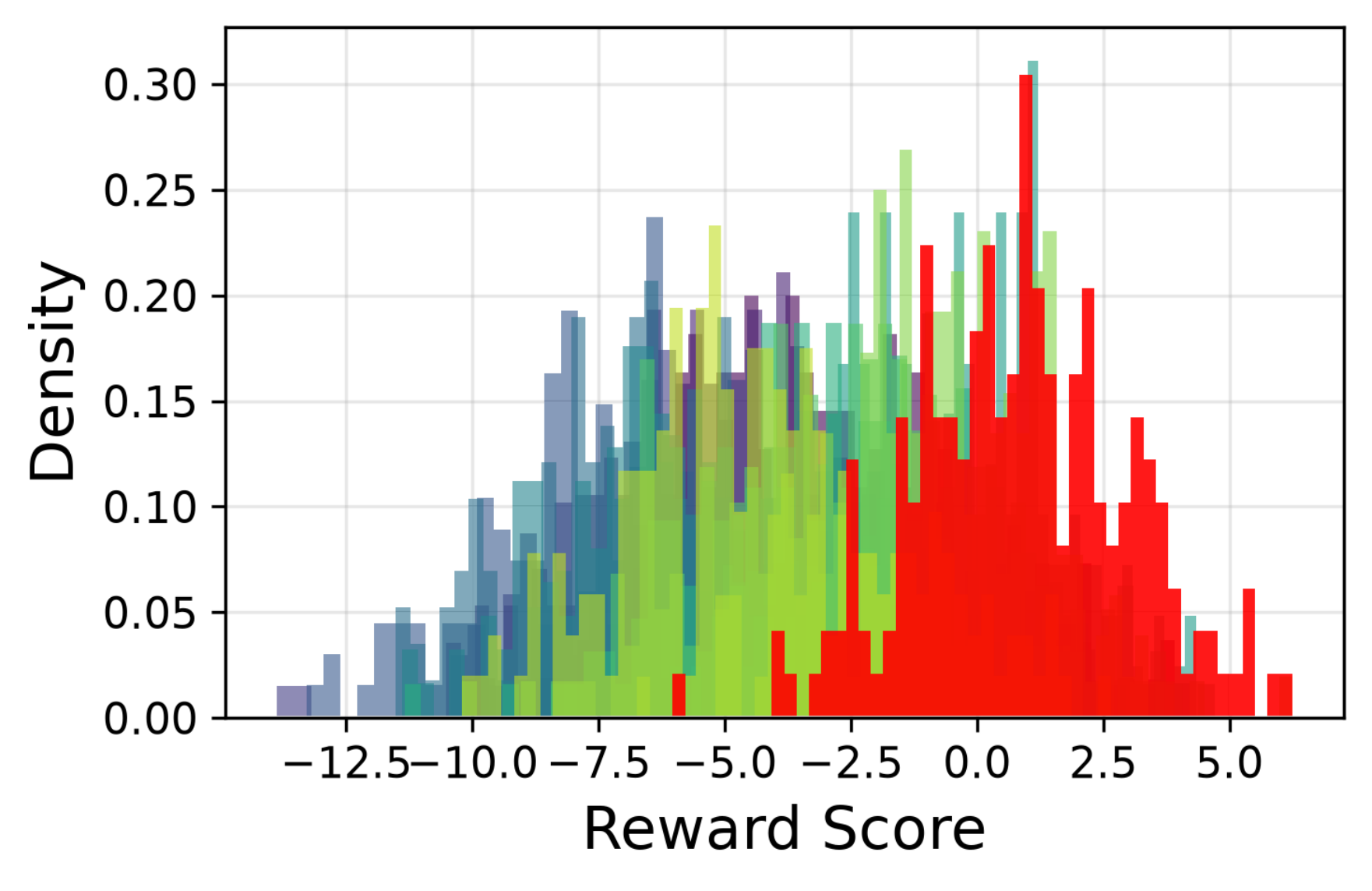}
    \\(b) After stage ($\alpha=0.5$)
  \end{minipage}
  \hfill
  \begin{minipage}[t]{0.32\linewidth}
    \centering
    \includegraphics[width=\linewidth]{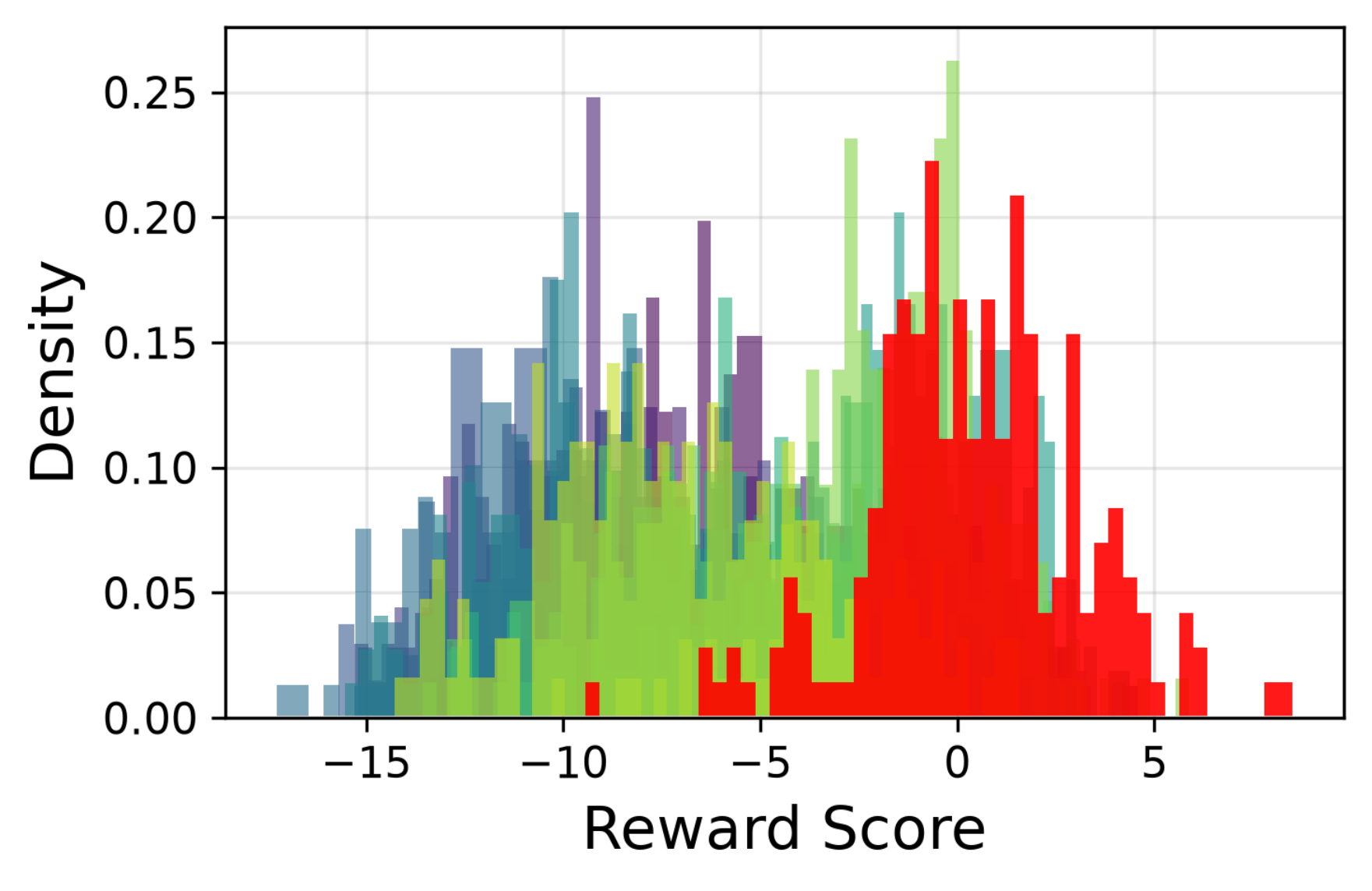}
    \\(c) After 2nd stage ($\alpha=0.75$)
  \end{minipage}
  
\caption{\textbf{Score distributions of the detector.} The detector is trained to identify GPT-4o as the target LLM. \textbf{\textcolor{red}{Red}} indicates the scores of the target model.}

\vspace{-0.1in}
  \label{fig:gpt_distribution}
\end{figure*}

\begin{figure*}[t]
\vspace{-0.1in}
  \centering
  \begin{minipage}[t]{0.32\linewidth}
    \centering
    \includegraphics[width=\linewidth]{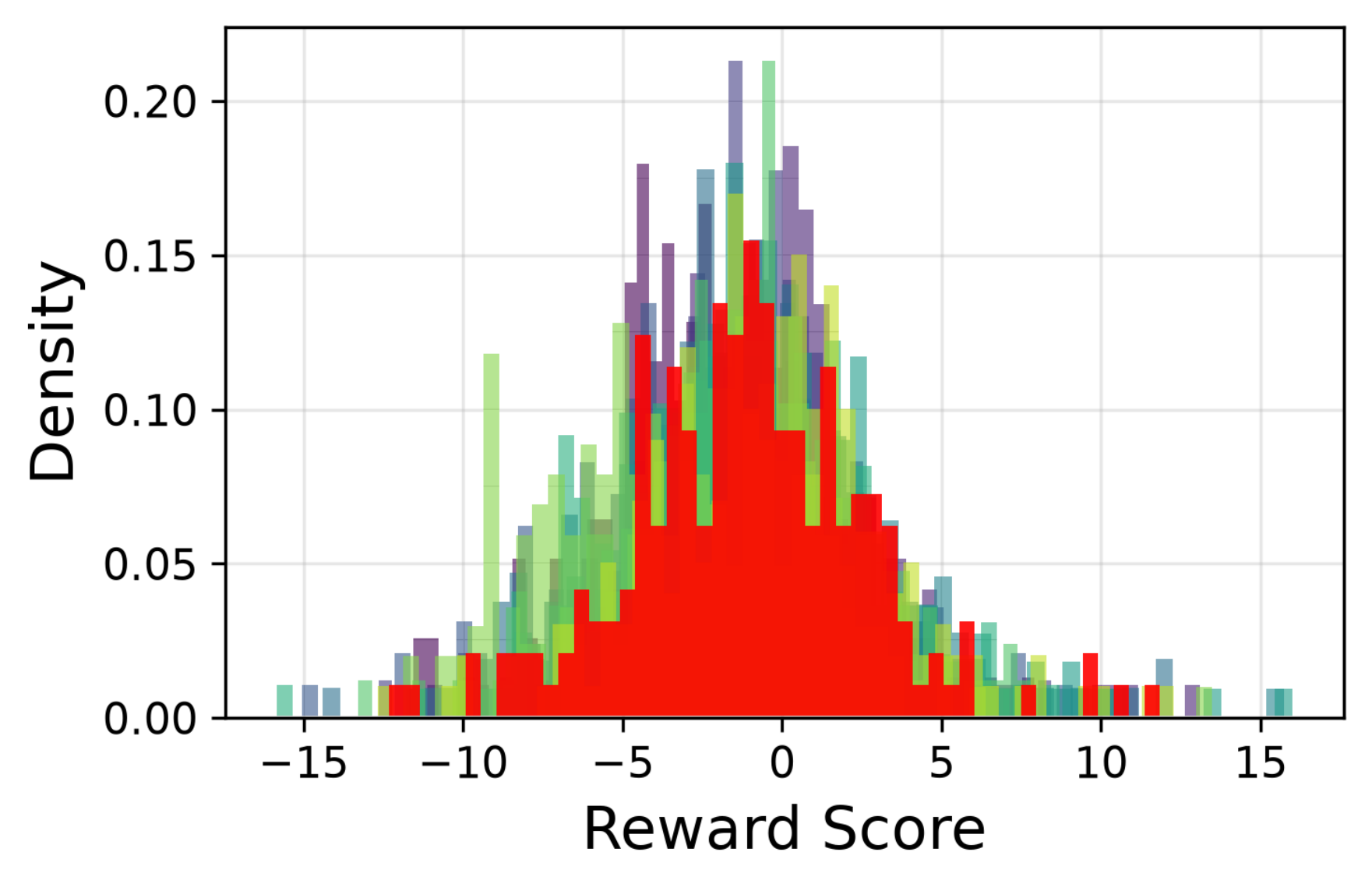}
    \\(a) Initial detector
  \end{minipage}
  \hfill
  \begin{minipage}[t]{0.32\linewidth}
    \centering
    \includegraphics[width=\linewidth]{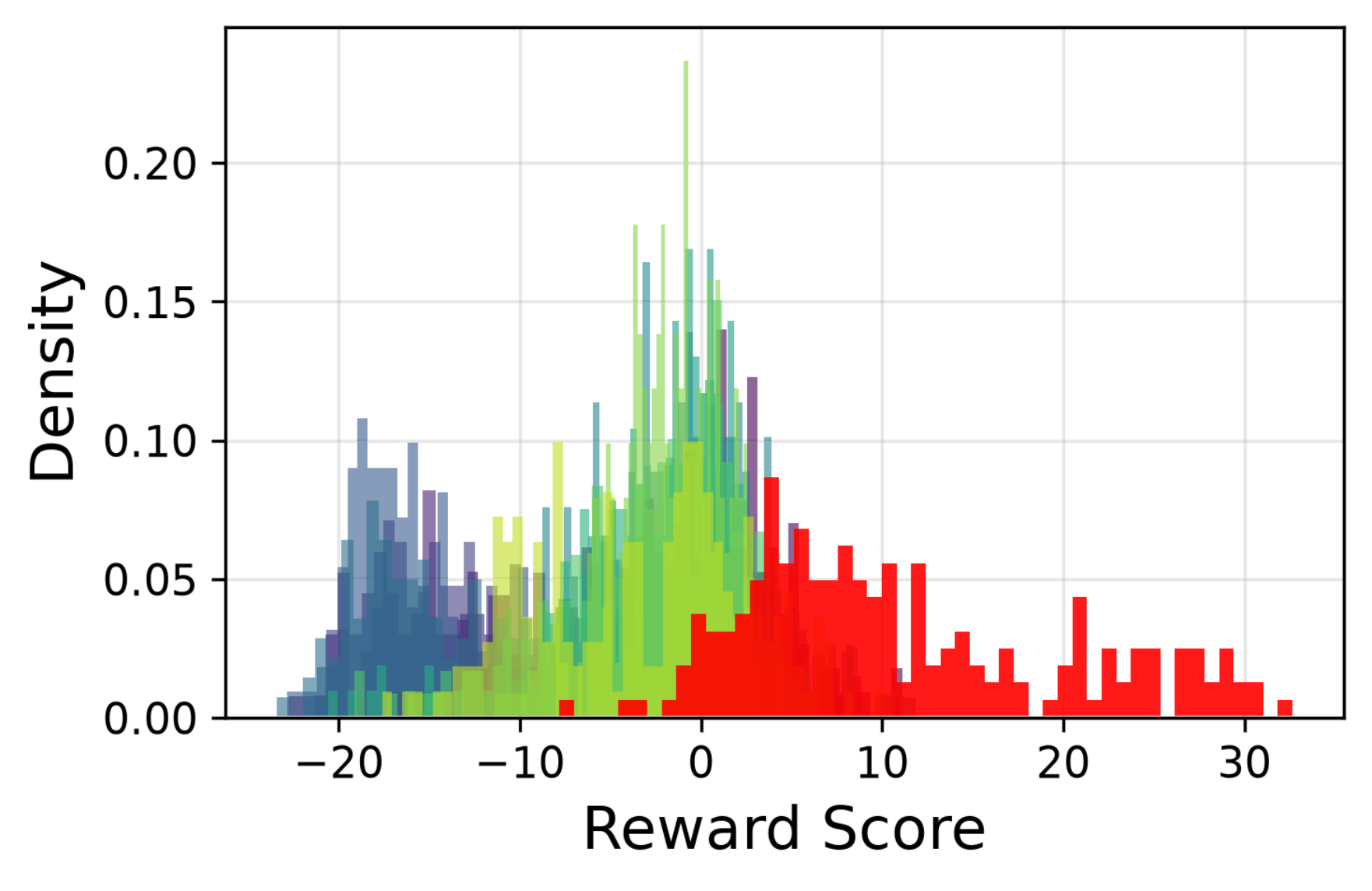}
    \\(b) After 1st stage ($\alpha=0.5$)
  \end{minipage}
  \hfill
  \begin{minipage}[t]{0.32\linewidth}
    \centering
    \includegraphics[width=\linewidth]{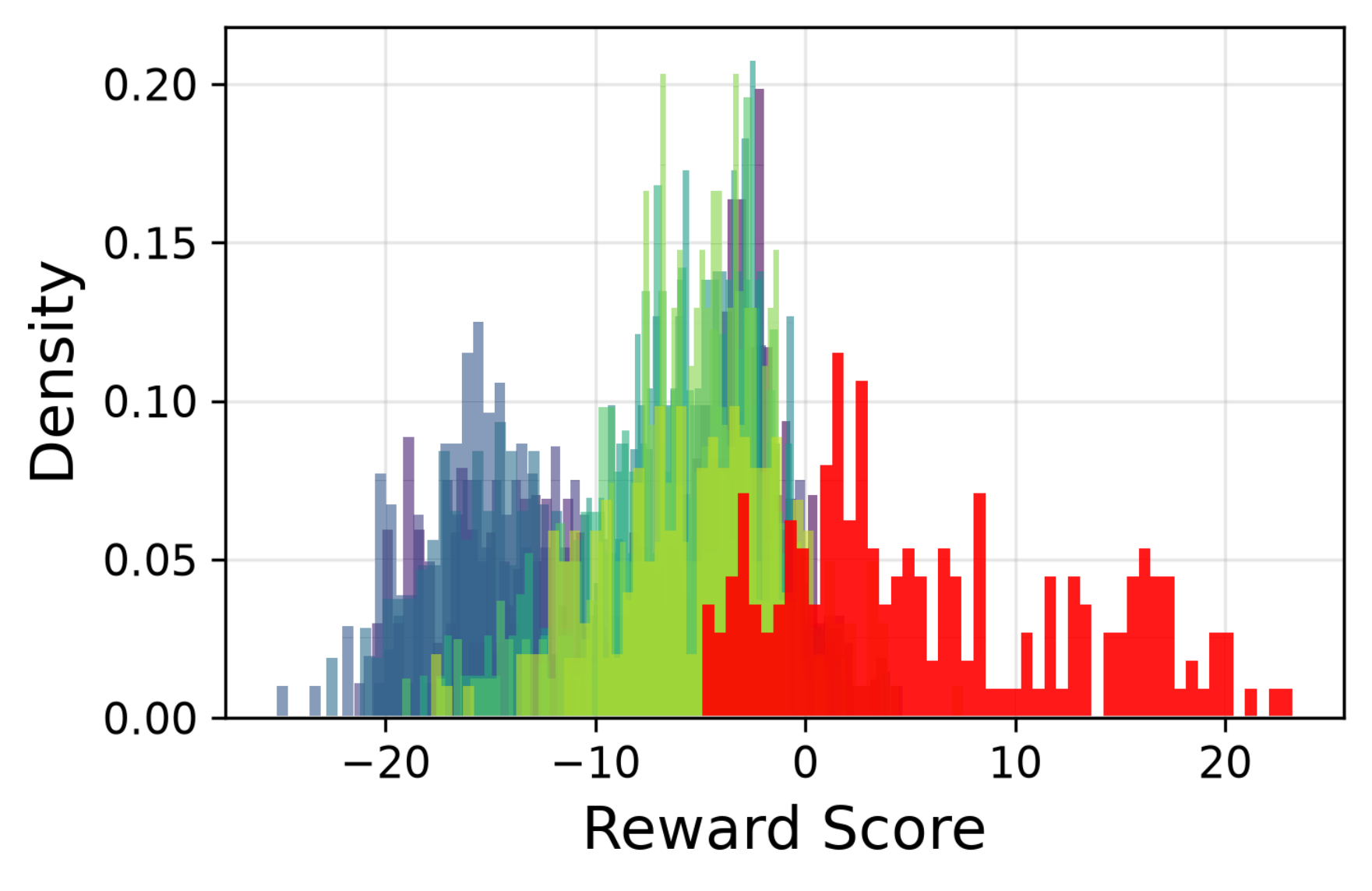}
    \\(c) After 2nd stage ($\alpha=0.75$)
  \end{minipage}
  
\caption{\textbf{Score distributions of the detector.} The detector is trained to identify Sonnet-4 as the target LLM. \textbf{\textcolor{red}{Red}} indicates the scores of the target model.}

\vspace{-0.1in}
\label{fig:claude_distribution}
\end{figure*}


\begin{figure}[t]
  \centering
  \includegraphics[width=0.85\linewidth]{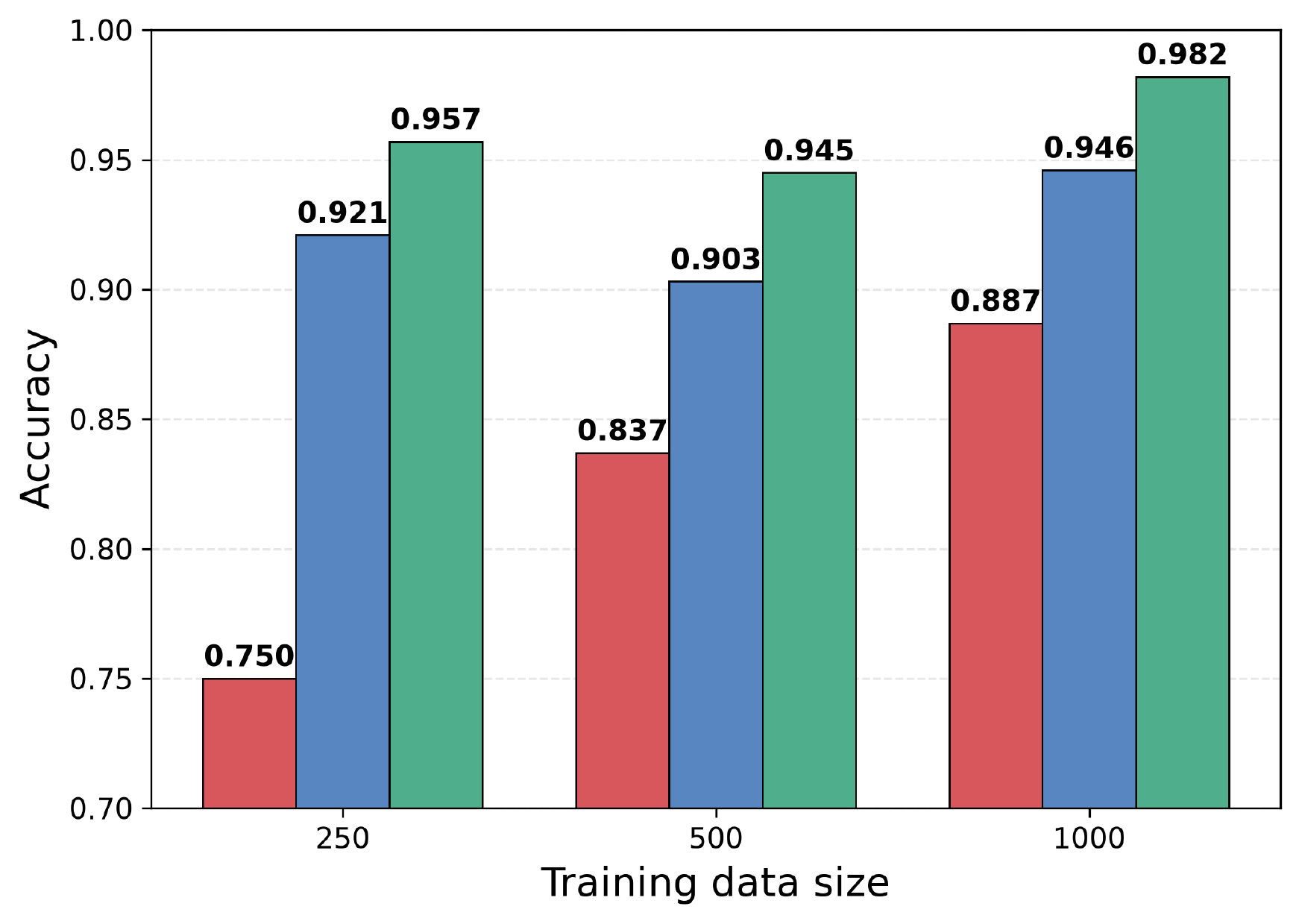}
    \caption{\textbf{Data scalability of \name{}.} Performance trend with the proposed training framework as the number of training samples increases from 250 to 1000. 
    Accuracy shows a generally linear improvement with larger training sets across all three target LLMs.
    Red bars represent GPT-4o,blue bars represent Gemini-Pro, and green bars represent Sonnet-4.}
  \label{fig:data_scalability_accuracy}
\end{figure}

\begin{table*}[t]
\small
\centering
\caption{\textbf{Effect of copy-model capacity on identification performance.}
We vary the capacity of Qwen3-based copy models from 0.6B to 8B parameters on the Alpaca dataset.
Increasing capacity from 0.6B to 4B improves overall performance, while further scaling to 8B provides no consistent additional gain across targets and metrics, suggesting that effective copy-modeling does not require a large-capacity backbone.}

\label{tab:rebuttal_copy_capacity}
\begin{tabular}{@{}lcccccc@{}}
\toprule
\multirow{2}{*}{\textbf{Copy Model}} 
    & \multicolumn{2}{c}{\textbf{GPT-4o}} 
    & \multicolumn{2}{c}{\textbf{Gemini-Pro}} 
    & \multicolumn{2}{c}{\textbf{Sonnet-4}} \\
\cmidrule(lr){2-3} \cmidrule(lr){4-5} \cmidrule(lr){6-7}
 & Acc. & AUROC & Acc. & AUROC & Acc. & AUROC \\
\midrule
Qwen3-0.6B & 0.826 & 0.819 & 0.941 & 0.947 & 0.938 & 0.942 \\
Qwen3-4B   & 0.839 & 0.845 & 0.963 & 0.964 & 0.991 & 0.953 \\
Qwen3-8B   & 0.878 & 0.824 & 0.949 & 0.963 & 0.954 & 0.955 \\
\bottomrule
\end{tabular}
\vspace{-0.1in}
\end{table*}

\begin{table}[t]
\centering
\small
\caption{\textbf{Candidate-pool generalization.}
The detector is trained using only seven similar-Elo competitors and evaluated on seen competitors, unseen outsider models, and the full composed pool.
The results show that \name{} generalizes beyond a fixed training-time candidate set.}
\vspace{0.1in}
\begin{tabular}{lccc}
\toprule
\textbf{Evaluation split} & \textbf{\# of Models} & {Acc.} & {AUROC} \\
\midrule
Seen competitors & 7 & 0.929 & 0.958 \\
Unseen outsiders & 9 & 0.854 & 0.814 \\
All composed & 16 & 0.887 & 0.884 \\
\bottomrule
\end{tabular}
\label{tab:candidate_pool_generalization}
\vspace{-0.1in}
\end{table}
\begin{table}[t]
\centering
\small
\caption{
\textbf{Cross-source transfer between Alpaca and Arena}
We train the detector on one query source and evaluate it on the other. Strong transfer performance indicates that \name{} does not merely exploit source-specific prompt artifacts.
}
\label{tab:cross_source_transfer}
\resizebox{\columnwidth}{!}{%
\begin{tabular}{lcccc}
\toprule
\multirow{2}{*}{\textbf{Target}} 
& \multicolumn{2}{c}{\textbf{Alpaca $\rightarrow$ Arena}} 
& \multicolumn{2}{c}{\textbf{Arena $\rightarrow$ Alpaca}} \\
\cmidrule(lr){2-3} \cmidrule(lr){4-5}
& Acc. & AUROC & Acc. & AUROC \\
\midrule
GPT-4o      & 0.868 & 0.838 & 0.883 & 0.861 \\
Gemini-Pro  & 0.940 & 0.976 & 0.928 & 0.956 \\
Sonnet-4    & 0.922 & 0.937 & 0.896 & 0.918 \\
\bottomrule
\end{tabular}
}
\end{table}

In this section, we further present score distribution of the detector for GPT-4o and Sonnet-4, similar to the case of Gemini-Pro in Figure~\ref{fig:gemini_distribution}.
The results of GPT-4o and Sonnet-4 are presented in Figures \ref{fig:gpt_distribution} and \ref{fig:claude_distribution}, respectively.

As shown in Figure~\ref{fig:gpt_distribution}, (a) the initial detector has no capability to distinguish the target model from the others, as their score distributions almost completely overlap. 
After the first triplet training stage with interpolation factor $\alpha=0.5$ (b), the distribution of the target model (red) shifts toward the positive side, indicating that the detector begins to assign systematically higher scores to the target responses. 
Meanwhile, the non-target models’ scores are pushed leftward, with their average moving toward approximately $-5.0$. 
Finally, under iterative curriculum training with $\alpha=0.75$ (c), the separation becomes much clearer: the target distribution concentrates further on the positive axis, while the non-target models form a distinct peak around $-10$, reflecting the detector’s strengthened discriminative capability.

Moreover, as shown in Figure~\ref{fig:claude_distribution}, (a) the initial detector again has no ability to distinguish the target model from the others, with the score distributions almost completely overlapping. 
After the first triplet training stage with interpolation factor $\alpha=0.5$ (b), the detector starts to separate certain models such as the blue one, while the distinction between the target model (red) and another non-target (green) remains ambiguous. 
However, following iterative curriculum training with $\alpha=0.75$ (c), the red, green, and blue distributions become much more clearly separated, demonstrating a substantially enhanced detection capability. 

Taken together, examining the score distributions across three different target models shows that \name{} progressively stregthens the detector with stronger discriminative power, while emphasizing once more that such gains are achieved through nearly cost-free synthetic data generation.

\subsection{Data scalability}
\label{sec:data_scalability}

Figure~\ref{fig:data_scalability_accuracy} examines how detection performance scales with the number of training data in the Alpaca dataset. 
It is observed that Accuracy shows a generally increasing trend as the number of training samples grows from 250 to 1000, suggesting that additional data could further enhance performance. 
Nevertheless, with only 1000 instructions which corresponds to a modest API collection cost, detector using \name{} already achieves strong results (average Accuracy above 0.918 across three target LLMs). 
This demonstrates that \name{} does not rely on prohibitively large datasets: even with a few hundred samples the detector remains competitive, and with 1000 samples it already provides robust accuracy in distinguishing state-of-the-art models.

\subsection{Effect of Copy-Model Capacity on Identification Performance}
\label{Effect of Copy-Model Capacity on Identification Performance}

In Table~\ref{tab:rebuttal_copy_capacity}, we analyze the impact of copy-model capacity on identification performance by varying the size of the Qwen3 backbone from 0.6B to 8B parameters \citep{qwen3}.

For these Qwen3-based experiments, all parameter-interpolated variants tended to emit the special token ``\texttt{<think>}'' at the beginning of generations. To ensure comparability across models, we strip these scaffolding tokens and retain only the final response for both training and evaluation.

We keep the evaluation setup identical to Section~\ref{sec:Experiment_setup}.
Increasing the copy-model capacity from 0.6B to 4B improves overall identification performance, whereas further scaling to 8B provides no consistent additional gain across targets and metrics.
This trend is consistent with the role of the copy model, which is used to approximate target-specific response characteristics for hard-negative synthesis rather than to maximize general-purpose language-modeling capability. These results suggest that a moderate-capacity copy model is sufficient for effective hard-negative generation, and that simply increasing copy-model size does not necessarily translate into better identification performance.

\subsection{Generalization beyond Training-Time Conditions}
\label{sec:generalization}

We next ask whether this performance depends on fixed training-time conditions.
In particular, we examine three possible sources of overfitting: the set of non-target models observed during training ,the query source used to train the detector, and the target model's decoding configuration at test time.
If \name{} merely memorizes a fixed candidate pool, exploits source-specific prompt artifacts, or relies on a particular decoding setting, its performance should degrade substantially when these conditions change.

\paragraph{Candidate-pool generalization.}
We first evaluate whether the detector depends on a fixed training-time competitor set.
For each target, we train the detector using only the seven close-Elo competitors and evaluate it on three splits: the seen competitors, nine unseen outsider models excluded from training, and the full composed pool of 16 non-target models.

As shown in Table~\ref{tab:candidate_pool_generalization}, \name{} achieves the strongest performance on the seen close-Elo competitors, but also generalizes to unseen outsider models excluded from the training-time competitor pool.
On the full composed pool, \name{} maintains 0.887 Accuracy and 0.884 AUROC on average across the three target LLMs.
These results indicate that \name{} is not simply learning a fixed lookup over predefined competitors.
Instead, it captures target-discriminative response patterns that transfer beyond the training-time candidate pool.

\paragraph{Cross-source transfer.}
We further test whether \name{} relies on source-specific prompt artifacts.
To do so, we train the detector on one query source and evaluate it on the other: Alpaca$\rightarrow$Arena and Arena$\rightarrow$Alpaca.

As shown in Table~\ref{tab:cross_source_transfer}, \name{} remains effective under both transfer directions.
For GPT-4o, which is the hardest target in our experiments, \name{} still achieves 0.868 Accuracy / 0.838 AUROC for Alpaca$\rightarrow$Arena and 0.883 Accuracy / 0.861 AUROC for Arena$\rightarrow$Alpaca.
For Gemini-Pro and Sonnet-4, transfer performance is even stronger, with AUROC values above 0.91 across both directions.

These results demonstrate that the attribution signal learned by \name{} is source-robust: it transfers across substantially different query distributions and reflects model-specific response patterns rather than dataset-specific artifacts.

\paragraph{Robustness to decoding configurations.}
We further examine whether \name{} remains effective when the target model's decoding configuration differs from that used to construct the training corpus.
In our main experiments, API-model responses are generated using greedy decoding.
We therefore regenerate only the 200 Gemini-Pro test responses using temperature $0.7$ and top-$p=0.9$, while keeping the trained detector and the responses from the eleven non-target models fixed.
This results in 2,200 target--opponent test pairs and requires no detector retraining.

Under greedy decoding, \name{} achieves 0.946 Accuracy and 0.967 AUROC.
With temperature $0.7$ and top-$p=0.9$, performance remains nearly unchanged at 0.949 Accuracy and 0.964 AUROC.
Thus, changing the target-side decoding configuration results in only a $+0.26$ percentage-point change in Accuracy and a $-0.29$ percentage-point change in AUROC.
These results indicate that the learned target-specific response signatures are robust to moderate changes in decoding and are not tied to the greedy-decoding configuration used during data collection.

\subsection{Additional Interpolation-Trajectory Validation}
\label{sec:additional_interpolation_validation}

\begin{table}[t]
\centering
\small
\caption{
\textbf{Validation of the interpolation trajectory across target models}.
Target-response perplexity (PPL) decreases as $\alpha$ increases toward the target-mimicking copy endpoint, while generation-level embedding similarity (Emb. Sim.) increases consistently.
}
\label{tab:interpolation_validation}
\resizebox{\columnwidth}{!}{
\begin{tabular}{llcc}
\toprule
\textbf{Target} & \textbf{Model} & PPL $\downarrow$ & Emb. Sim. $\uparrow$ \\
\midrule
\multirow{5}{*}{GPT-4o}
& Base (Phi-4)      & 2.114 & -- \\
& $\alpha=0.25$   & 1.940 & 0.795 \\
& $\alpha=0.50$   & 1.813 & 0.802 \\
& $\alpha=0.75$   & 1.724 & 0.815 \\
& Copy            & 1.668 & -- \\
\midrule
\multirow{5}{*}{Gemini-Pro}
& Base (Phi-4)      & 4.190 & -- \\
& $\alpha=0.25$   & 3.811 & 0.693 \\
& $\alpha=0.50$   & 3.516 & 0.707 \\
& $\alpha=0.75$   & 3.306 & 0.727 \\
& Copy            & 3.187 & -- \\
\midrule
\multirow{5}{*}{Sonnet-4}
& Base (Phi-4)      & 3.977 & -- \\
& $\alpha=0.25$   & 3.494 & 0.746 \\
& $\alpha=0.50$   & 3.125 & 0.758 \\
& $\alpha=0.75$   & 2.864 & 0.771 \\
& Copy            & 2.701 & -- \\
\bottomrule
\end{tabular}
}
\vspace{-0.2in}
\end{table}

Table~\ref{tab:interpolation_validation} extends the
\textit{Validation of interpolation-based difficulty} analysis in Section~\ref{sec:analysis} to Gemini-Pro and Sonnet-4. 
Consistent with GPT-4o, target-response PPL decreases monotonically with $\alpha$, while embedding similarity to the target responses increases.
Thus, the interpolation ordering is consistent across all three target LLMs.

\subsection{Extended Evaluation on GPT-5}

\begin{table}[t]
\centering
\small
\caption{\textbf{Extended identification evaluation on GPT-5.}
We evaluate target-present identification with GPT-5 as the target model under the same setting as the main experiments. Best results are highlighted in \textbf{bold}.}
\begin{tabular}{lcc}
\toprule
\textbf{Method} & {Acc.} & {AUROC} \\
\midrule
Len-W & 0.448 & 0.381 \\
Len-C & 0.446 & 0.384 \\
TF--IDF & 0.821 & 0.904 \\
BoW & 0.797 & 0.906 \\
\cellcolor{gray!20}\name{}
& \cellcolor{gray!20}\textbf{0.960}
& \cellcolor{gray!20}\textbf{0.983} \\
\bottomrule
\end{tabular}
\label{tab:gpt5_results}
\end{table}

To examine whether \name{} remains effective for a more recent frontier model, we extend the target-present identification experiment to GPT-5\footnote{\texttt{gpt-5-2025-08-07}}.
Following the same evaluation protocol as Table~\ref{tab:main_result}, we replace GPT-4o with GPT-5 as the target model, collect responses for the same 1,400-query pool, and train a target-specific detector using the same Phi-4-mini copy-model backbone.
We then compare \name{} against the same statistical identification baselines.

As shown in Table~\ref{tab:gpt5_results}, \name{} achieves 0.960 Accuracy and 0.983 AUROC.
The strongest statistical baseline reaches 0.821 Accuracy with TF--IDF and 0.906 AUROC with BoW, corresponding to improvements of 13.91 and 7.64 percentage points, respectively.
These results indicate that \name{} remains effective for GPT-5 despite the substantial capability gap between the frontier target and the lightweight copy-model backbone.

While a fixed 4B-scale copy backbone may not be equally suitable for every future frontier model, \name{} is model-agnostic and can use stronger open-source copy backbones as target capability increases.


\subsection{Qualitative Interpretability Analysis}
\label{sec:qualitative_interpretability}

To complement the aggregate erasure-based results in
Table~\ref{tab:interpretability_erasure}, we examine a representative GPT-4o example for which \name{} correctly identifies the target response while TF--IDF incorrectly favors the non-target response.

We consider the prompt \texttt{``Describe the benefits of virtual reality in healthcare.''}
For this example, \name{} correctly identifies the GPT-4o response with a discrimination margin of 17.16, whereas TF--IDF incorrectly favors the opponent response.
The individual words receiving high importance under \name{} include ordinary explanatory terms such as ``improve,'' ``adherence,'' ``treatment,'' ``enhancing,'' ``outcomes,'' ``skills,'' ``education,''
and ``healthcare.''

To determine whether these cues are merely isolated lexical markers or belong to a coherent semantic span, we additionally measure the effect of removing each sentence from the response.
The sentence causing the largest reduction in the \name{} margin is:
\texttt{``Virtual reality (VR) is increasingly being integrated into healthcare,
offering a range of benefits that enhance both patient care and medical
training.''}

Among the 354 words in the response, words belonging to this sentence receive an average importance rank of 138.43 under \name{}, compared with 174.29 under TF--IDF and 194.07 under BoW, where a lower rank
indicates greater importance. Thus, \name{} assigns substantially greater importance to the same
explanatory span than either lexical baseline.
Together with the aggregate erasure results in Table~\ref{tab:interpretability_erasure}, this example suggests that
\name{} captures not only isolated word-frequency patterns but also
semantically coherent explanatory phrasing characteristic of the target
model's responses.

\subsection{Cost Analysis}
\label{sec:cost analysis}
As training detector with \name{} incurs the additional cost, one might ask whether it is more effective than simply collecting more target-specific responses.
Regarding this, we note that additional target-model API collection and \name{} serve complementary roles: API scaling increases the number of real target responses available for supervision, whereas \name{} extracts more structured and discriminative supervision from the responses already collected through interpolated hard negatives.
Nevertheless, under a fixed resource budget, it is useful to compare their marginal cost and performance gains.

As shown in Figure~\ref{fig:data_scalability_accuracy}, increasing the training set from 250 to 500 responses for each of the three targets requires 750 additional API generations, costing approximately \$2.44, while improving average Accuracy from 0.876 to 0.895 (+0.019).
In contrast, Table~\ref{tab:ablation_study} shows that the full \name{} configuration improves average Accuracy from 0.863 to 0.951 (+0.088) under a fixed real-response budget.
The additional computation required by \name{}---including copy-model fine-tuning, interpolated-response generation, and one additional detector-training stage---takes approximately 51.5 GPU minutes per target, corresponding to about \$1.14 at \$1.33 per GPU-hour.
Thus, under comparable marginal costs, \name{} yields substantially larger gains by extracting richer supervision from the available target responses rather than merely increasing the number of API-collected examples.

\section{Usage of AI assistants}

In preparing this work, we used AI-based writing assistants to improve sentence structure, correct grammatical errors, and enhance overall readability. These tools were employed soley for language refinement and did not contribute to the development of technical content, research methodology, or experimental analysis. All scientific ideas, results, and conclusions presented in paper were conceived and authored entirely by the researchers. The use of AI assistance was restricted to editorial purposes and did not affect the originality or intellectual contributions of the work.

\end{document}